\documentclass{article}

\usepackage[preprint]{neurips_2026}

\usepackage[utf8]{inputenc} % allow utf-8 input
\usepackage[T1]{fontenc}    % use 8-bit T1 fonts
\usepackage{hyperref}       % hyperlinks
\usepackage{url}            % simple URL typesetting
\usepackage{booktabs}       % professional-quality tables
\usepackage{amsfonts}       % blackboard math symbols
\usepackage{nicefrac}       % compact symbols for 1/2, etc.
\usepackage{microtype}      % microtypography
\usepackage{xcolor}         % colors
\usepackage{graphicx}
\usepackage{amsmath}
\usepackage{multirow}
\usepackage{enumitem}
\usepackage{wrapfig}
\usepackage{algorithm}

\usepackage{amssymb}
\usepackage{algpseudocode}

\title{DynT2I-Eval: A Dynamic Evaluation Framework for Text-to-Image Models}

\author{%
  Juntong Wang$^1$, \quad
  Jiarui Wang$^1$, \quad
  Huiyu Duan$^{1}$, \quad
  Lewei Li$^1$, \\ % 删掉了末尾多余的 \quad
  \textbf{Guangtao Zhai$^{1}$, \quad Xiongkuo Min$^{1}$} \\
  \\
  $^1$Institute of Image Communication and Network Engineering, \\
  Shanghai Jiao Tong University, Shanghai, China \\
}

\begin{document}

\maketitle

\begin{abstract}
Existing text-to-image (T2I) benchmarks largely rely on fixed prompt sets, leaving them vulnerable to overfitting and benchmark contamination once publicly released and repeatedly reused. In this work, we propose \textbf{DynT2I-Eval}, a fully automated \underline{dyn}amic \underline{eval}uation framework for \underline{T2I} models. It constructs a structured visual semantic space from long-form descriptions, decomposing prompts into controllable dimensions (\textit{e.g.}, subject, logical constraint, environment, and composition). This enables the continuous generation of fresh prompts via task-specific spaces and difficulty-aware sampling. DynT2I-Eval evaluates model performance across text alignment, perceptual quality, and aesthetics. Heterogeneous outputs are unified into prompt-conditioned pairwise comparisons, allowing a dynamic scheduler, micro-batch aggregation, and weighted Bayesian updates to maintain a stable online leaderboard despite changing prompt distributions and model injection. Experiments with independently sampled prompt streams demonstrate that continually refreshed prompts provide a robust evaluation protocol, reducing the impact of prompt-set-specific tuning. Simulations and ablations further confirm that the proposed ranking framework achieves a strong balance among cold-start convergence, late-entry discovery, and long-run ranking fidelity.
\end{abstract}

\section{Introduction}
Recent advances in text-to-image (T2I) generation \cite{zimage,flux-2,flux1,flux1kreadev2025,sd3} have substantially improved model capabilities in semantic rendering, perceptual fidelity \cite{visualqualityr1,qinsight,wang2025lmm4lmm}, and aesthetic expression \cite{artimuse,cao2025unipercept}. As model development accelerates, reliable evaluation has become increasingly important for model comparison, system selection, and research progress. However, most existing T2I benchmarks \cite{DPG-Bench,wang2025lmm4lmm,tit-score,T2i-compbench++,TIIF-Bench,genai-bench,GenEval2} are built on fixed prompt sets, where models are compared on a predefined collection of prompts. While such static benchmarks have played an important role in early-stage evaluation, their limitations are becoming increasingly pronounced once the prompt sets are publicly released and reused. In particular, model development may increasingly susceptible to explicit or implicit optimization toward specific test distributions, causing leaderboard performance to drift away from true practical open-world generalization ability. These limitations motivate the need for a more scalable and contamination-resistant evaluation paradigm beyond static prompt-set benchmarking.

\begin{figure}
    \centering
    \includegraphics[width=1\linewidth]{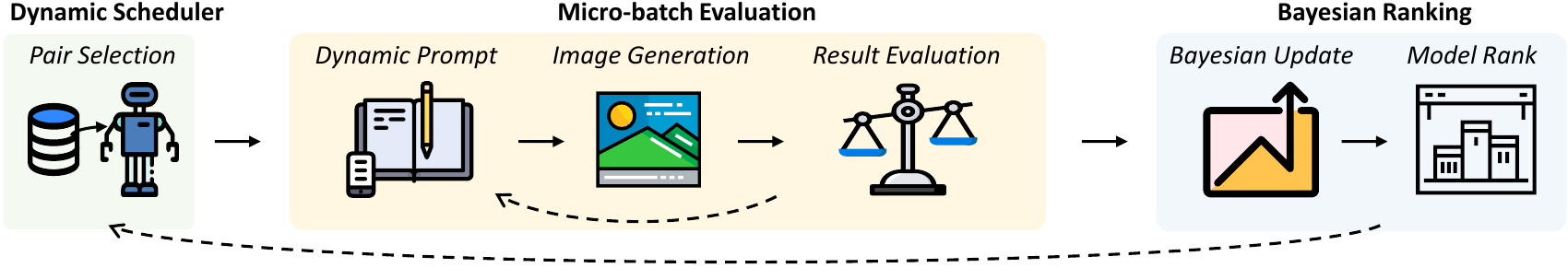}
    \vspace{-12pt}
    \caption{Overview of the DynT2I-Eval framework. The system continuously generates dynamic prompts and performs micro-batch evaluations via prompt-conditioned pairwise comparison. A dynamic scheduler selects model pairs to balance global coverage and local rank resolution. The evaluation outcomes are aggregated using Bayesian updates to maintain an online, continually evolving leaderboard.}
    \label{fig_main}
    \vspace{-12pt}
\end{figure}

A natural direction is to move toward dynamic evaluation, where benchmarks continuously introduce fresh prompts over time rather than relying on a fixed and repeatedly exposed prompt set. In principle, such a paradigm can mitigate prompt-set overfitting and make evaluation better aligned with open-world deployment. However, dynamic evaluation for T2I generation is not as simple as repeatedly replacing prompts. First, dynamically generated prompts must exhibit sufficient quality and discriminative power, \textit{i.e.}, they should cover diverse semantic and visual demands while meaningfully separating model capabilities, rather than degenerating into random, trivial, or ambiguous test cases. Second, once the prompt set changes across evaluation rounds, the scores and rankings of the models no longer reside within a shared reference frame, making cross-round alignment and stable comparison inherently challenging. A practical dynamic evaluation framework for T2I models must therefore jointly address two coupled problems, \textit{i.e.}, continuously producing high-quality evaluation prompts and maintaining reliable ranking under a changing prompt distribution.

To this end, we propose \textbf{DynT2I-Eval}, a dynamic evaluation framework for text-to-image models as illustrated in Figure~\ref{fig_main}. DynT2I-Eval replaces one-shot static testing with a continual process of prompt generation \cite{OnoeDocci2024,qwen3.5}, pairwise comparison \cite{huang2024vbench,i2i-bench}, and online update \cite{trueskill,trueskill2,Glicko-2,ELO,Active-Ranking,Rank-Centrality,RUCB}. Specifically, the framework first constructs a structured visual semantic space from long-form image descriptions \cite{OnoeDocci2024}, and then designs task-specific prompt spaces and dynamic sampling mechanisms for three dimensions, including text alignment \cite{qwen3.5}, perceptual quality \cite{qinsight,wang2025lmm4lmm,visualqualityr1}, and aesthetic quality \cite{artimuse,cao2025unipercept}. This allows the benchmark to continually produce fresh test samples with diversity, judgeability, and discriminative value. During evaluation, heterogeneous evaluator results across these dimensions are converted into prompt-conditioned pairwise combinations for the same prompt. Built on this interface, the system further performs online scheduling to select informative model pairs, and combines micro-batch aggregation with weighted Bayesian posterior updates to maintain a stable and updatable leaderboard under the conditions of changing prompt distributions and continuous model injection. Compared with fixed prompt-set benchmarks, this design reduces the risk of targeted optimization after benchmark exposure and allows evaluation benchmark itself to continuously evolve alongside rapid model advancements.

Our contributions are summarized as follows:
\begin{itemize}[leftmargin=*]
    \item We introduce a dynamic evaluation paradigm for text-to-image models that replaces fixed prompt-set benchmarking with the evaluation based on dynamically generated prompts, mitigating prompt-set overfitting caused by targeted optimization after benchmark exposure and enabling evaluation to evolve alongside model progress.
    \item We develop a structured dynamic prompt engine that extracts a controllable visual semantic space from long-form image descriptions and constructs task-specific prompt spaces and sampling mechanisms from three dimensions including text alignment, perceptual quality, and aesthetic quality, thereby enabling the continual creation of test prompts with diversity, judgeability, and discriminability.
    \item We propose an online scheduling and Bayesian ranking framework for dynamic leaderboards. The framework unifies heterogeneous evaluators into a prompt-conditioned pairwise interface, employs conservative uncertainty-penalized scores for leaderboard ranking while using mean-driven criteria for model scheduling, and combines micro-batch aggregation with weighted posterior updates to achieve stable ranking under changing prompts and continual model injection.
\end{itemize}

\section{Related Work}
\subsection{Static T2I Benchmarks}
Existing text-to-image benchmarks \cite{SpatialGenEval,saneval,compalign} advance fine-grained evaluation but inherently suffer from fixed prompt distributions. Representative frameworks incorporate dual-track evaluation in T2I-CompBench++ \cite{T2i-compbench++}, attribute binding in GenAI-Bench \cite{genai-bench}, and atomic assessments in DPG-Bench \cite{DPG-Bench} and GenEval 2 \cite{GenEval2}. Other notable efforts include DreamBench++ \cite{dreambench++} for personalized generation and TIIF-Bench \cite{TIIF-Bench} for difficulty-stratified prompts. Despite these advances, most existing benchmarks are built on fixed prompt sets. As models are developed with awareness of these benchmarks, leaderboard performance may be affected by prompt exposure, benchmark-specific tuning, or implicit optimization toward the fixed prompt set. Our work addresses this limitation by introducing a dynamic prompt generator that creates a nearly infinite prompt space through orthogonal dimension decomposition.

\subsection{Dynamic Evaluation}
To overcome these static limitations, dynamic evaluation emerges as a promising direction. Early explorations in the visual generation domain such as DyEval \cite{DyEval} propose interactive frameworks based on user feedback, yet their reliance on human participation severely restricts scalability and automation. Conversely, fully automated dynamic frameworks demonstrate robust contamination-resistant capabilities in large language model assessments, exemplified by SQUID GAME \cite{Squid-Game}, MACEval \cite{Maceval}, Prism, and LLMEval-Fair \cite{LLMEval-Fair}. To our knowledge, fully automated dynamic evaluation for T2I models remains underexplored. We therefore develop a framework that combines structured prompt-space construction with online Bayesian ranking, enabling scalable dynamic evaluation without requiring human-in-the-loop comparisons during leaderboard updates.

\subsection{Rank System}
Reliable model comparison within this dynamic paradigm necessitates robust skill rating systems. The Elo \cite{ELO} algorithm estimates participant strength based on pairwise match results but assumes static skills and lacks uncertainty quantification. Subsequent methods address these deficiencies, where Glicko-2 \cite{Glicko-2} models confidence using rating deviation, TrueSkill \cite{trueskill} extends Bayesian inference to team scenarios via factor graphs, and K-Sort Arena \cite{k-sort} introduces multi-dimensional comparisons for efficient ranking. Since these systems primarily process static competition data, we adapt Bayesian skill inference techniques to suit an environment where models continuously face evolving prompts. This adaptation establishes a solid theoretical foundation for reliable model comparison under non-stationary conditions.

\begin{figure}
    \centering
    \includegraphics[width=1\linewidth]{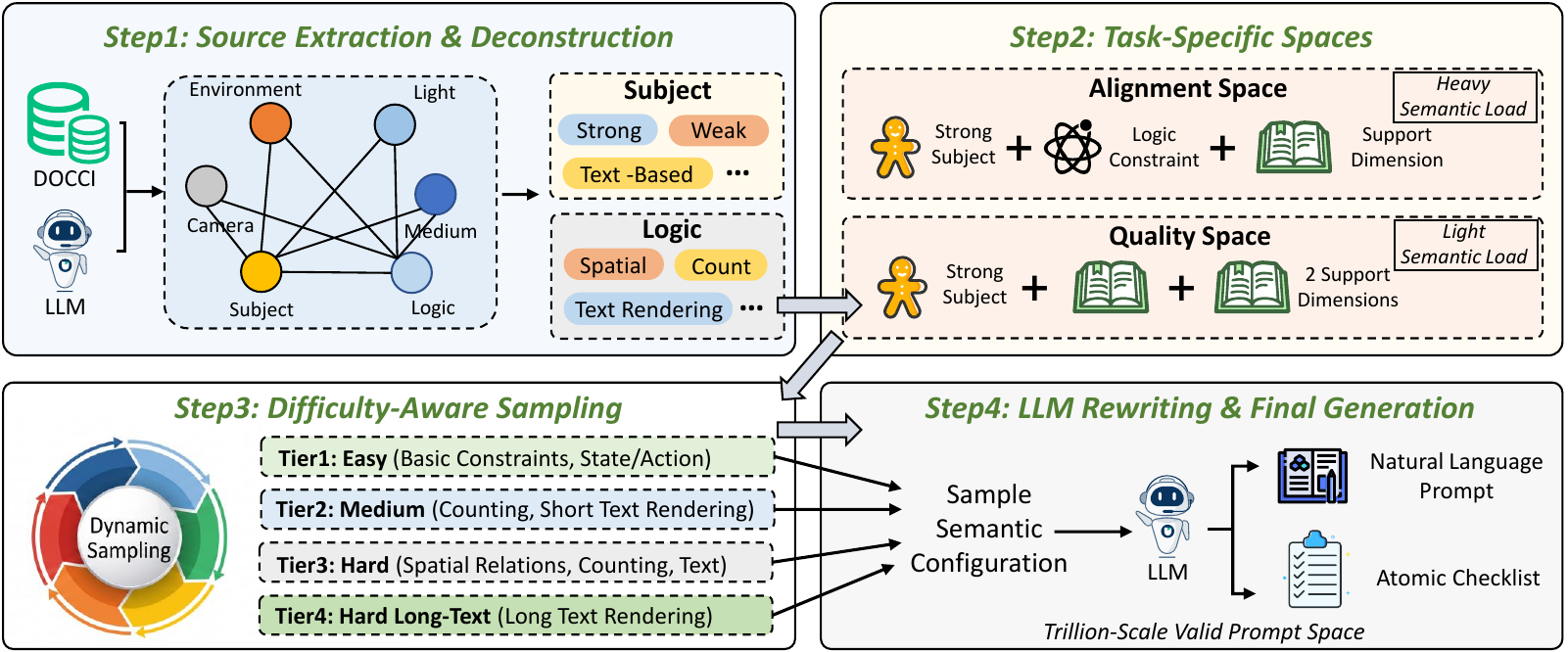}
    \vspace{-12pt}
    \caption{The dynamic prompt generation pipeline. Long-form image descriptions are first deconstructed into structured semantic dimensions (\textit{e.g.}, subject, environment, logic). The system then constructs task-specific prompt spaces with different semantic loads for text alignment and quality evaluation. Finally, semantic configurations are sampled under a difficulty-aware policy (Tier 1 to 4) and rewritten by an LLM into natural language prompts and atomic checklists.}
    \label{prompt}
    \vspace{-10pt}
\end{figure}

\section{Dynamic Prompt Generator}

\subsection{Structured Prompt Construction from Long-form Descriptions}
As depicted in Figure~\ref{prompt}, our prompt construction pipeline starts from 14,845 long-form image descriptions collected from DOCCI \cite{OnoeDocci2024}. Rather than directly rewriting them into prompts, we decompose each description into six semantic dimensions: subject, cognitive logic constraints, environment, lighting and atmosphere, camera and composition, and medium or format. This decomposition converts dense natural-language descriptions into controllable semantic units and serves as a controllable source for prompt construction. We further reorganize the decomposed units according to their evaluability and sampling utility. In particular, subject candidates are filtered to retain visually salient and self-contained entities. Logic constraints are regrouped into categories, including attribute binding, counting, spatial relations, action or state understanding, and text rendering. This process turns long-form descriptions into a structured prompt source that supports controllable sampling, difficulty adjustment, and task-specific evaluation.

\subsection{Task-specific Prompt Spaces}
We construct separate prompt spaces for alignment evaluation \cite{qwen3.5} and quality evaluation \cite{qinsight,visualqualityr1,artimuse,cao2025unipercept,AIGCIQA2023}, since the two tasks require different prompt distributions. The alignment space is designed to evaluate semantic faithfulness, and each prompt combines a clear subject, an explicit logic constraint, and one supporting dimension selected from environment, lighting and atmosphere, camera and composition, or medium and format. This keeps prompts visually grounded while preserving strong and judgeable constraints. In contrast, the quality space is designed for perceptual and aesthetic assessment, where overly complex semantics can distract evaluators from image quality. Therefore, quality prompts use lighter compositions, typically combining one strong subject with two supporting dimensions, while introducing logic constraints only with low probability. This task-specific separation improves both controllability and sampling efficiency.

\subsection{ Difficulty-aware Dynamic Sampling}
To balance diverse and controllable, prompts are generated online under a difficulty-aware sampling policy rather than predefined as a fixed list. This design is particularly important for alignment evaluation, where challenge should come from deliberate control over constraint types rather than accidental variation in wording.
The alignment space is organized into four difficulty modes: easy, medium, hard, and hard long-text. Easy prompts focus on relatively basic constraints such as attribute binding and state or action understanding. Medium prompts add counting and short text rendering, increasing semantic load without becoming overly specialized. Hard prompts emphasize more on spatial relations, together with counting and short text rendering, which better differentiate stronger text-to-image models. The hard long-text mode is reserved for long text rendering, a capability that remains challenging enough to merit a dedicated high-difficulty setting.
For each sample, the system first draws a semantic configuration from the corresponding task space and then rewrites it into natural language with an LLM \cite{qwen3.5}. Alignment prompts additionally come with an atomic checklist derived from the same sampled structure, so the prompt and its evaluation targets remain aligned. Since sampling follows a fixed construction policy, the prompt stream can continue to expand without drifting away from the intended evaluation distribution. We define the prompt space as the number of rule-valid structured configurations before LLM rewriting. Under the current candidate pools and validity constraints, this space is already on the order of $10^{12}$ for both alignment and quality, and becomes substantially larger when optional logic combinations are expanded. Compared with a benchmark built from a frozen prompt list, this design offers broader coverage and reduces the effectiveness of prompt-level memorization.

\begin{figure}
    \centering
    \includegraphics[width=1\linewidth]{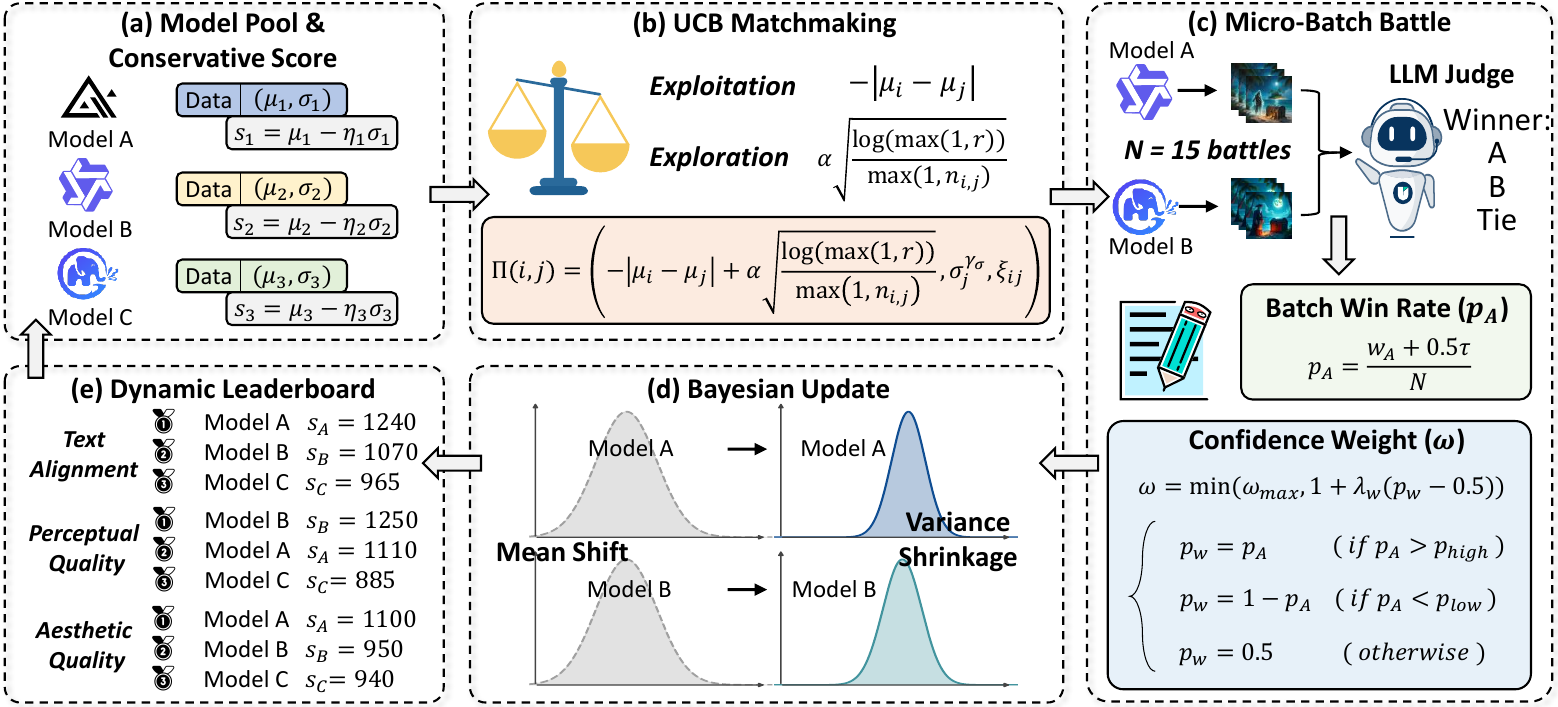}
    \vspace{-13pt}
    \caption{Overview of the dynamic ranking and scheduling framework. The system selects candidate model pairs using a UCB-style matchmaking strategy to balance exploration (uncertainty) and exploitation (posterior mean closeness). Pairwise comparisons are conducted in micro-batches. The micro-batch outcomes are weighted by batch-level confidence and used for Bayesian updates that shifts the posterior mean and shrinks the variance, resulting in a conservative and stable dynamic leaderboard.}
    \label{fig:Third figure}
    \vspace{-9pt}
\end{figure}

\section{Dynamic Evaluation and Ranking}

\subsection{Multi-dimensional Evaluators}

To improve leaderboard interpretability, we decompose text-to-image evaluation into three independent dimensions: text alignment, perceptual quality, and aesthetic quality. Each dimension maintains its own leaderboard rather than being collapsed into a single mixed score. Although the three dimensions use different evaluators, their outputs are all mapped into a unified prompt-conditioned pairwise format \cite{huang2024vbench} before entering the downstream ranking module.
For text alignment, we use a Qwen3.5-based vision-language \cite{qwen3.5,gu2025surveyllmasajudge} judge that operates directly in pairwise form. The judge is conditioned on both the original prompt and its atomic checklist, and is instructed to focus only on whether visible constraints are satisfied, while ignoring unrelated factors such as beauty or stylistic preference. To reduce order bias, each comparison is performed twice with swapped input order, and the final decision is aggregated into an outcome in $\{A,B,\mathrm{Tie}\}$.
For perceptual quality, we use VisualQuality-R1 \cite{visualqualityr1}, a SOTA IQA model \cite{visualqualityr1,qinsight,NIMA,NIQE,BRISQUE,ke2021musiq,CLIPIQA,yang2022maniqa,C2Score,qalign,DeQA}. Following its uncertainty-aware evaluation protocol, we draw $K$ stochastic samples for each image and characterize perceptual quality with the resulting mean and variance. When comparing two images under the same prompt, we convert these statistics into a relative win probability using a Thurstone model, and then discretize it into $\{A,B,\mathrm{Tie}\}$ with a predefined tie region. This is more robust than comparing a single score directly.
For aesthetic quality, we use ArtiMuse \cite{artimuse}, a SOTA IAA model \cite{artimuse,cao2025unipercept}. This dimension compares scalar aesthetic scores directly and applies a tie threshold so that marginal cases are not forced into overly confident win-loss decisions.
Overall, although the three dimensions rely on different evaluator types, they are all mapped into a unified prompt-conditioned pairwise outcome: direct pairwise judgment for alignment, uncertainty-aware probabilistic comparison for perceptual quality, and scalar-score comparison with a tie threshold for aesthetic quality. This unified interface serves as the basis for the downstream dynamic ranking scheduler.

\subsection{Dynamic Ranking Framework}

\subsubsection{Ranking State and Pair Selection}

As shown in Figure~\ref{fig:Third figure}, each model $i$ is associated with a Bayesian rating state $(\mu_i,\sigma_i)$, where $\mu_i$ denotes the posterior mean and $\sigma_i$ denotes the posterior uncertainty. For leaderboard display, we use a conservative score
\begin{equation}
s_i=\mu_i-\eta_i \sigma_i,
\end{equation}
where the uncertainty penalty is warmed up for newly injected models as
\begin{equation}
\eta_i=\eta \cdot \mathrm{clip}\!\left(\frac{c_i}{C_{\mathrm{warm}}},\, r_{\min},\, 1\right).
\end{equation}
Here $c_i$ is the number of accumulated valid comparisons for model $i$, $C_{\mathrm{warm}}$ is the warmup horizon, and $r_{\min}$ is the minimum penalty ratio. This preserves conservative leaderboard display while avoiding excessive early suppression of late-entry models.

At scheduling round $r$, the scheduler first selects as pivot the active model with the fewest accumulated comparisons, ensuring balanced global coverage. It then selects an opponent using a UCB-style criterion by maximizing the lexicographic tuple
\begin{equation}
\Pi(i,j)=\left(
-|\mu_i-\mu_j|+\alpha\sqrt{\frac{\log(\max(1,r))}{\max(1,n_{ij})}},
\ \sigma_j^{\gamma_\sigma},
\ \xi_{ij}
\right),
\end{equation}
where $n_{ij}$ is the number of previous comparisons between models $i$ and $j$, $\alpha$ controls exploration strength, $\gamma_\sigma$ controls the uncertainty preference in tie-breaking, and $\xi_{ij}$ is a random jitter used only as a last-resort tie-breaker. The first term favors locally informative comparisons between models with similar posterior means, while the second term allocates more budget to under-sampled pairs. In this way, leaderboard display is driven by the conservative score $s_i$, whereas pair scheduling is driven by posterior means for efficient local rank refinement.

\subsubsection{Micro-Batch Aggregation and Confidence-weighted Update}

Each selected model pair is evaluated on a micro-batch of prompts rather than a single prompt. Suppose model $A$ obtains $w_A$ wins, model $B$ obtains $w_B$ wins, and the number of ties is $\tau$, yielding $N=w_A+w_B+\tau$ valid comparisons. We summarize the batch by the empirical winning rate
\begin{equation}
p_A=\frac{w_A+0.5\tau}{N}.
\end{equation}
The batch is then converted into a ternary macro outcome using two decisiveness thresholds:
\begin{equation}
\mathrm{macro}(A,B)=
\begin{cases}
A, & p_A>p_{\mathrm{high}},\\
B, & p_A<p_{\mathrm{low}},\\
\mathrm{Tie}, & \text{otherwise}.
\end{cases}
\end{equation}

To quantify batch-level evidence strength, we define a confidence weight
\begin{equation}
\omega=\min\!\left(\omega_{\max},\,1+\lambda_w(p_w-0.5)\right),
\end{equation}
where
\begin{equation}
p_w=
\begin{cases}
p_A, & p_A>p_{\mathrm{high}},\\
1-p_A, & p_A<p_{\mathrm{low}},\\
0.5, & \text{otherwise}.
\end{cases}
\end{equation}
Thus, tie batches receive unit weight, whereas more decisive batches induce larger updates up to the cap $\omega_{\max}$.

For decisive batches, we apply a standard Gaussian-approximate Bayesian pairwise update to the winner and loser, using $\omega$ to scale the mean update magnitude and to moderately strengthen uncertainty shrinkage. Since this step follows the standard update used in Bayesian rating systems, we omit its closed-form expressions. For tie batches, instead of discarding the observation, we apply a fixed soft tie update:
\begin{equation}
\mu_A'=(1-\rho)\mu_A+\rho\frac{\mu_A+\mu_B}{2},\qquad
\mu_B'=(1-\rho)\mu_B+\rho\frac{\mu_A+\mu_B}{2},
\end{equation}
while both posterior standard deviations are multiplicatively shrunk, subject to a lower bound. This design suppresses prompt-level noise, preserves informative near-equal outcomes, and yields stable online ranking under a limited comparison budget. More details can be found in Appendix.

In all experiments, we use the fixed configuration $(p_{\mathrm{low}}, p_{\mathrm{high}})=(0.42,0.58)$, $\omega_{\max}=2.0$, $\lambda_w=2.0$, $\rho=0.05$, and a tie-side uncertainty shrinkage factor of $0.98$. These hyperparameters were selected by grid search on simulated environments covering different skill gaps, noise levels, and tie rates, and then frozen for all real benchmark experiments without benchmark-specific or model-specific retuning.

\begin{table*}[t]
\centering
\scriptsize
\caption{Main dynamic leaderboard at round 250 (taking $\sim$48 hours per dimension). Rankings are reported separately for text-image alignment, perceptual quality, and aesthetic quality. For each dimension, \textbf{Rk} denotes the rank, $s=\mu-3\sigma$ is the conservative score (following the standard $3\sigma$ rule), and $\mu,\sigma$ are the posterior mean and uncertainty, respectively. Models marked with $*$ are late-entry anchors injected after the initial leaderboard had largely stabilized.}
\label{tab:main_leaderboard}
\resizebox{\textwidth}{!}{
\begin{tabular}{lccccccccccccc}
\toprule
\multirow{2}{*}{\textbf{Model}} 
& \multicolumn{4}{c}{\textbf{Text Alignment}} 
& \multicolumn{4}{c}{\textbf{Perceptual Quality}} 
& \multicolumn{4}{c}{\textbf{Aesthetic Quality}} \\
\cmidrule(lr){2-5} \cmidrule(lr){6-9} \cmidrule(lr){10-13}
& \textbf{Rk} & \textbf{$s$} & \textbf{$\mu$} & \textbf{$\sigma$}
& \textbf{Rk} & \textbf{$s$} & \textbf{$\mu$} & \textbf{$\sigma$}
& \textbf{Rk} & \textbf{$s$} & \textbf{$\mu$} & \textbf{$\sigma$} \\
\midrule
FLUX.2-klein-9B \cite{flux-2}      & 1  & 1507.69 & 1617.64 & 36.65 & 4  & 1059.62 & 1142.07 & 27.49 & 9  & 565.02  & 653.12  & 29.37 \\
LongCat-Image \cite{LongCat-Image}     & 2  & 1361.83 & 1458.22 & 32.13 & 2  & 1407.11 & 1532.49 & 41.80 & 3  & 1198.11 & 1315.04 & 38.98 \\
FLUX.2-klein-4B*  \cite{flux-2} & 3  & 1200.82 & 1368.95 & 56.04 & 8  & 883.49  & 1003.02 & 39.84 & 10 & 510.62  & 607.86  & 32.42 \\
FLUX.1-Krea-dev  \cite{flux1kreadev2025}  & 4  & 1133.53 & 1239.13 & 35.20 & 9  & 733.53  & 821.74  & 29.40 & 1  & 1470.65 & 1579.26 & 36.20 \\
Z-Image-Turbo  \cite{zimage}    & 5  & 1124.74 & 1232.08 & 35.78 & 1  & 1443.63 & 1576.85 & 44.41 & 12 & 347.66  & 446.19  & 32.84 \\
SD-3.5-Large     \cite{sd3} & 6  & 829.00  & 948.65  & 39.88 & 11 & 571.97  & 658.61  & 28.88 & 4  & 1183.99 & 1308.78 & 41.60 \\
FLUX.1-dev    \cite{flux1}   & 7  & 771.57  & 868.78  & 32.40 & 3  & 1171.46 & 1288.40 & 38.98 & 5  & 949.07  & 1057.33 & 36.09 \\
SD-3.5-Medium    \cite{sd3}  & 8  & 763.20  & 862.19  & 32.99 & 10 & 613.03  & 690.86  & 25.94 & 7  & 714.93  & 835.18  & 40.09 \\
CogView4-6B  \cite{zheng2024cogview}  & 9  & 567.48  & 662.80  & 31.77 & 12 & 567.45  & 651.46  & 28.00 & 11 & 462.32  & 555.52  & 31.07 \\
SD-3-Medium  \cite{sd3} & 10 & 550.83  & 637.25  & 28.81 & 7  & 905.71  & 1001.90 & 32.06 & 8  & 691.53  & 813.23  & 40.56 \\
SD3.5-Large-Turbo  \cite{sd3}  & 11 & 483.74  & 599.04  & 38.43 & 6  & 994.74  & 1073.75 & 26.33 & 6  & 945.98  & 1051.20 & 35.07 \\
SDXL-Base-1.0* \cite{podell2023sdxl}  & 12 & -48.15  & 246.50  & 98.22 & 5  & 1019.97 & 1121.73 & 33.92 & 2  & 1405.49 & 1517.68 & 37.40 \\
\bottomrule
\end{tabular}
}
\vspace{-10pt}
\end{table*}

\section{Experiments}
\subsection{Main Dynamic Leaderboard and Insights}

We evaluate 12 text-to-image models in the dynamic arena and report three separate leaderboards for text-image alignment, perceptual quality, and aesthetic quality. All comparisons use a micro-batch size of 15, and two models (SDXL-Base-1.0 \cite{podell2023sdxl} and FLUX.2-klein-4B \cite{flux-2}) are injected after the initial leaderboard has largely stabilized to test late-entry positioning.

As shown in Table~\ref{tab:main_leaderboard}, the three leaderboards differ substantially. FLUX.2-klein-9B \cite{flux-2} leads the alignment ranking, Z-Image-Turbo performs best in perceptual quality, and FLUX.1-Krea-dev \cite{flux1kreadev2025} achieves the strongest aesthetic ranking, while SDXL-Base-1.0 \cite{podell2023sdxl} remains weak in alignment but ranks highly in aesthetics. These discrepancies indicate that current T2I models exhibit clear capability trade-offs, and that alignment, perceptual fidelity, and aesthetics should be evaluated separately rather than merged into a single score.

\begin{wraptable}{r}{0.65\textwidth}
\centering
\scriptsize
\caption{Human validation of automatic evaluators on 450 sampled pairwise comparison cases. Relaxed agreement counts borderline \textit{Tie}--vs.--\textit{A/B} cases as consistent, while strict agreement requires an exact ternary match.}
\label{tab:human_validation}
\setlength{\tabcolsep}{3.5pt}
\begin{tabular}{lrrrrrrrr}
\toprule
\textbf{Dim.} & \textbf{N} & \textbf{Relaxed} & \textbf{Strict} & \textbf{$\kappa$} & \textbf{W-$\kappa$} & \textbf{P} & \textbf{S} & \textbf{K} \\
\midrule
Alignment          & 150 & 94.67\% & 84.67\% & 0.704 & 0.702 & 0.704 & 0.702 & 0.686 \\
Perceptual quality & 150 & 89.33\% & 81.33\% & 0.665 & 0.693 & 0.695 & 0.694 & 0.672 \\
Aesthetic quality  & 150 & 92.67\% & 76.00\% & 0.603 & 0.679 & 0.681 & 0.681 & 0.648 \\
\midrule
Overall            & 450 & 92.22\% & 80.66\% & 0.662 & 0.696 & 0.696 & 0.695 & 0.671 \\
\bottomrule
\end{tabular}
\vspace{-10pt}
\end{wraptable}

The ranking is already largely stable by round 250, with later variations mainly occurring among closely matched models. Moreover, the late-entry anchors are rapidly localized near their final ranking regions of the leaderboard, showing that the proposed dynamic ranking framework can efficiently incorporate new models. We also note that the aesthetic leaderboard should be interpreted more cautiously, since aesthetic judgment is inherently subjective and strongly conditioned on the evaluator and its training distribution. In practice, the marginal cost of adding a new model is low, as scores typically stabilize within about 30 rounds, or roughly 450 generated images per model.

In addition, we conduct a small-scale human validation of the automatic evaluators for the three dimensions. We randomly sample 150 pairwise comparison cases for each dimension---alignment, perceptual quality, and aesthetic quality---resulting in 450 cases in total. Each case contains a prompt, an image pair, and the automatic evaluator decision, and human annotators assign a ternary label from $\{A,B,\mathrm{Tie}\}$. Because the boundary between a weak preference and a tie is often subjective, we report both relaxed and strict agreement. Relaxed agreement counts borderline \textit{Tie}--vs.--\textit{A/B} cases as consistent, whereas strict agreement requires an exact ternary match. As shown in Table~\ref{tab:human_validation}, the overall relaxed and strict agreement rates are 92.22\% and 80.66\%, respectively, with $\kappa=0.662$ and weighted $\kappa=0.696$. The correlation statistics are also high ($P=0.696$, $S=0.695$, $K=0.671$), indicating substantial agreement between the automatic evaluators and human judgments. Overall, these results support the reliability of the evaluators for large-scale dynamic evaluation, while confirming that borderline \textit{Tie} cases remain the main source of disagreement.

% 请确保在文档导言区（\begin{document} 之前）添加这行代码：
% \usepackage{wrapfig}

\subsection{Reduced Transferability of Prompt-Set-Specific Optimization}
\label{sec:anti_targeted_optimization}

A central advantage of dynamic evaluation is that leaderboard performance is measured on continually refreshed prompts rather than on a single exposed prompt set. This makes evaluation less dependent on prompt-set-specific tuning and better aligned with generalizable model behavior. We simulate this setting by independently sampling four 500-prompt sets, A/B/C/D, from the same generator, using A as the exposed tuning set and B/C/D as private dynamic streams. To avoid conflating this effect with our own ranking pipeline, we use two external evaluators, VQAScore~\cite{vqascore} and LMM4LMM~\cite{wang2025lmm4lmm}, and tune only inference hyperparameters.

For SD3.5 Medium~\cite{sd3} and FLUX.2-klein-4B~\cite{flux-2}, we search 26 and 24 configurations, respectively, select top configurations on A, and test them on B/C/D. Table~\ref{tab:targeted_opt_transfer} reports their transfer ratios and private-set ranks, showing weak transfer across independently sampled prompt sets. For SD3.5 Medium \cite{sd3}, the A-selected top configuration drops to average private ranks of 22.0/26 and 19.3/26 under VQAScore \cite{vqascore} and LMM4LMM \cite{wang2025lmm4lmm}, respectively. For FLUX.2-klein-4B \cite{flux-2}, selected configurations remain relatively high-ranked, but their transfer ratios are near zero or negative across private streams, indicating that the gains found on A do not reliably transfer. Thus, dynamic prompt streams reduce the influence of prompt-set-specific tuning on leaderboard outcomes and provides a more robust evaluation protocol.

\begin{table}[t]
\centering
\caption{Transfer of top configurations selected on prompt set A to independently sampled prompt sets B/C/D for SD3.5 Medium~\cite{sd3} and FLUX.2-klein-4B~\cite{flux-2}. Values are reported in the order B/C/D. The transfer ratio is defined as the gain over the default configuration on a private set divided by the gain over the default configuration on A. Ranks are computed among all candidate configurations, and lower rank is better.}
\label{tab:targeted_opt_transfer}
\scriptsize
\setlength{\tabcolsep}{2.0pt}
\renewcommand{\arraystretch}{1.0}
\resizebox{\columnwidth}{!}{
\begin{tabular}{llcccc}
\toprule
\multirow{2}{*}{\textbf{Evaluator}} 
& \multirow{2}{*}{\textbf{Metric}} 
& \multicolumn{2}{c}{\textbf{SD3.5 Medium \cite{sd3}}} 
& \multicolumn{2}{c}{\textbf{FLUX.2-klein-4B \cite{flux-2}}} \\
\cmidrule(lr){3-4}\cmidrule(lr){5-6}
& 
& \textbf{\shortstack{Top-1 on A}} 
& \textbf{Avg. Top-5 on A}
& \textbf{\shortstack{Top-1 on A}} 
& \textbf{Avg. Top-3 on A} \\
\midrule
\multirow{2}{*}{VQAScore \cite{vqascore}}
& TR on B/C/D 
& $-0.11 / 0.14 / 0.14$ 
& $0.12 / 0.32 / 0.37$ 
& $-0.0023 / -0.0002 / -0.0018$ 
& $-0.0024 / 0.0001 / -0.0018$ \\
& Rank on B/C/D 
& $23/26 \; / \; 21/26 \; / \; 22/26$ 
& $13/26 \; / \; 12.6/26 \; / \; 12.2/26$ 
& $3/24 \; / \; 6/24 \; / \; 4/24$ 
& $3.0/24 \; / \; 5.3/24 \; / \; 4.0/24$ \\
\midrule
\multirow{2}{*}{LMM4LMM \cite{wang2025lmm4lmm}}
& TR on B/C/D 
& $-0.14 / 0.06 / 0.12$ 
& $0.08 / 0.16 / 0.23$ 
& $-0.0028 / -0.0010 / -0.0004$ 
& $-0.0029 / -0.0011 / -0.0008$ \\
& Rank on B/C/D 
& $24/26 \; / \; 19/26 \; / \; 15/26$ 
& $14/26 \; / \; 11/26 \; / \; 10/26$ 
& $4/24 \; / \; 5/24 \; / \; 4/24$ 
& $3.7/24 \; / \; 4.7/24 \; / \; 4.3/24$ \\
\bottomrule
\end{tabular}
}
\vspace{-12pt}
\end{table}
\vspace{-5pt}

\begin{table*}[t]
\centering
\small
\caption{Multi-environment Monte Carlo results. We report Ours, K-Sort Arena~\cite{k-sort}, and the strongest remaining baseline (Best Other, selected per environment/metric) for SRCC at 500/1500/3000 rounds and late-entry discovery. Higher SRCC and lower discovery batches are better.}
\label{tab:simulation_results}
\resizebox{\textwidth}{!}{
\begin{tabular}{lcccccccccccc}
\toprule
\multirow{2}{*}{\textbf{Environment}}
& \multicolumn{3}{c}{\textbf{SRCC @ 500} $\uparrow$}
& \multicolumn{3}{c}{\textbf{SRCC @ 1500} $\uparrow$}
& \multicolumn{3}{c}{\textbf{SRCC @ 3000} $\uparrow$}
& \multicolumn{3}{c}{\textbf{Late-entry discovery} $\downarrow$} \\
\cmidrule(lr){2-4} \cmidrule(lr){5-7} \cmidrule(lr){8-10} \cmidrule(lr){11-13}
& \textbf{Ours} & \textbf{K-Sort} & \textbf{Best Other}
& \textbf{Ours} & \textbf{K-Sort} & \textbf{Best Other}
& \textbf{Ours} & \textbf{K-Sort} & \textbf{Best Other}
& \textbf{Ours} & \textbf{K-Sort} & \textbf{Best Other} \\
\midrule
reference & 0.8293 & 0.7474 & 0.6606 (RUCB) & 0.8845 & 0.8512 & 0.7602 (RUCB) & 0.9546 & 0.9283 & 0.8097 (Glicko-2) & 5.75 & 27.20 & 2.75 (AR) \\
noisier\_judge & 0.8236 & 0.6642 & 0.6883 (RUCB) & 0.9007 & 0.8358 & 0.7446 (RUCB) & 0.8912 & 0.9020 & 0.8042 (RUCB) & 5.75 & 26.75 & 2.75 (AR) \\
hard\_extreme & 0.8082 & 0.6263 & 0.6974 (RUCB) & 0.8809 & 0.8064 & 0.8251 (RUCB) & 0.8937 & 0.9366 & 0.8563 (RUCB) & 5.70 & 26.15 & 1.00 (AR) \\
less\_extreme & 0.8558 & 0.6798 & 0.6448 (RUCB) & 0.9376 & 0.9251 & 0.7527 (RUCB) & 0.8903 & 0.9240 & 0.7831 (Glicko-2) & 5.90 & 26.15 & 5.35 (AR) \\
prompt\_calibrated & 0.7970 & 0.6558 & 0.7335 (RUCB) & 0.9054 & 0.8477 & 0.7823 (RUCB) & 0.9104 & 0.8877 & 0.8327 (RUCB) & 5.80 & 26.30 & 2.90 (AR) \\
prompt\_clustered\_stress & 0.6029 & 0.5746 & 0.4875 (RUCB) & 0.6141 & 0.5944 & 0.5970 (RUCB) & 0.5969 & 0.6300 & 0.6212 (RUCB) & 6.05 & 26.45 & 1.05 (AR) \\
\bottomrule
\end{tabular}
}
\vspace{-8pt}
\end{table*}

\subsection{Ranking Robustness under Multi-environment Monte Carlo Simulation}
\label{sec:simulation}

To evaluate whether the proposed scheduler and Bayesian update mechanism can maintain accurate rankings, adapt to newly injected models, and remain robust under noisy or heterogeneous conditions, we construct a multi-environment Monte Carlo simulation that abstracts away the biases of specific multimodal evaluators while preserving key challenges of dynamic T2I benchmarking. We compare our method with a broad set of representative baselines, including ELO \cite{ELO}, TrueSkill \cite{trueskill}, TrueSkill2 \cite{trueskill2}, Glicko-2 \cite{Glicko-2}, Rank Centrality \cite{Rank-Centrality}, K-Sort Arena \cite{k-sort}, Active Ranking (AR) \cite{Active-Ranking}, and RUCB \cite{RUCB}. Each environment is run over multiple independent trials under the same comparison budget, and we report ranking quality at different stages together with late-entry discovery performance.

We consider six environments: \textit{reference} (base setting), \textit{noisier\_judge} (higher evaluator noise), \textit{hard\_extreme} and \textit{less\_extreme} (different levels of prompt difficulty), \textit{prompt\_calibrated} (prompt-level calibration shift), and \textit{prompt\_clustered\_stress} (clustered prompt perturbation under stress). These settings jointly vary evaluator noise, prompt difficulty, and prompt distribution perturbation, allowing us to test not only ranking accuracy in standard conditions but also robustness under stress. We report SRCC at 500, 1500, and 3000 rounds, as well as the average number of batches required to discover a newly injected strong model.

As shown in Table~\ref{tab:simulation_results}, our method achieves the best SRCC@500 and SRCC@1500 in all six environments, demonstrating consistently stronger early-stage ranking accuracy than all baselines. The advantage is especially pronounced in more challenging settings such as \textit{less\_extreme}, where our method substantially outperforms the strongest baseline at 500 rounds. At the same time, the discovery time of our method remains highly stable across environments, staying within 5.7--6.1 batches, and achieves a 100\% discovery success rate in all six cases. Reporting K-Sort Arena explicitly helps clarify the trade-off: K-Sort is the strongest long-run fidelity baseline and attains the best SRCC@3000 in some environments, but reacts much more slowly to newly injected models, whereas AR is the fastest specialist on late-entry discovery and RUCB is often the strongest remaining baseline in noisier or more perturbed settings. Overall, these results show that our scheduler provides a stronger efficiency--quality--robustness trade-off than existing ranking baselines for dynamic evaluation.

\subsection{Ablation Study of the Dynamic Ranking Policy}
We analyze the proposed ranking policy from two complementary perspectives: parameter calibration around the paper configuration, and removal of individual mechanisms.
Table~\ref{tab:param_ablation} compares the paper configuration against several nearby alternative settings under the unified six-environment dynamic ranking simulation. The paper configuration is used as the reference, and the other rows test nearby variants with different exploration strength, update aggressiveness, or warmup-related settings.
The reported Unified Obj.\ is the scalar criterion used to select the paper configuration, defined as the mean weighted score over the six environments, combining ranking accuracy, dynamic discovery, top-1 recovery, and baseline-relative win/margin terms. The paper configuration remains a strong operating point under this objective: although some alternatives improve individual metrics in specific cases, none provides a better overall balance across the main and stress environments. This supports our choice of the final policy as a robust balance rather than one optimized for a single metric.
Table~\ref{tab:module_ablation} examines mechanism contribution by removing key components from the same paper configuration. The reported Ours-only Obj.\ uses the same six-environment aggregation but removes the baseline-relative win/margin terms, since only Ours is re-evaluated; it therefore measures internal policy performance rather than margin over external baselines. Removing \textit{eta warmup} greatly slows the discovery of newly injected high-potential models, with Discovery degrading from 5.08 to 9.47, indicating that warmup mainly helps cold-start entrants surface early. Removing decisive confidence weighting causes the largest drop in the objective among the targeted removals and noticeably weakens early ranking quality, reducing Main SRCC@500 from 0.7803 to 0.6789. These observations indicate that the final policy benefits from both cold-start protection and confidence-aware updates, which together improve dynamic discovery and early convergence.

  \begin{table}[t]
  \centering
  \small
  \caption{Parameter comparison for selecting the final ranking policy under the unified six-environment simulation. Unified Obj.\ denotes the mean weighted six-environment
  score over ranking accuracy, dynamic discovery, top-1 recovery, and baseline-relative win/margin terms.}
  \label{tab:param_ablation}
  \resizebox{0.88\linewidth}{!}{%
  \begin{tabular}{lcccc}
  \toprule
  \textbf{Variant} & \textbf{Unified Obj.} & \textbf{$\Delta$ vs paper} & \textbf{Main SRCC@1500} & \textbf{Stress SRCC@1500} \\
  \midrule
  earlier warmup ref. ($\beta{=}75,\alpha{=}200$) & 5.4015 & -0.1721 & 0.8455 & 0.6102 \\
  more-explore ($\alpha{=}230,w{=}2.1$)            & 5.4480 & -0.1256 & \textbf{0.8678} & 0.6012 \\
  softer-updates ($\eta{=}2.8,w{=}1.8$)            & 5.3368 & -0.2368 & 0.8274 & \textbf{0.6939} \\
  paper config ($\beta{=}70,\alpha{=}190$)         & \textbf{5.5736} & \textbf{0.0000} & 0.8607 & 0.6118 \\
  \bottomrule
  \end{tabular}%
  }
  \vspace{-8pt}
  \end{table}

  \begin{table}[t]
  \centering
  \small
  \caption{Targeted mechanism ablation from the paper configuration. Ours-only Obj.\ uses the same six-environment aggregation but omits baseline-relative win/margin terms
  because only Ours is re-evaluated. Lower Discovery is better.}
  \label{tab:module_ablation}
  \resizebox{0.88\linewidth}{!}{%
  \begin{tabular}{lccccc}
  \toprule
  \textbf{Variant} & \textbf{Ours-only Obj.} & \textbf{$\Delta$ vs paper} & \textbf{Main SRCC@500} & \textbf{Stress SRCC@1500} & \textbf{Discovery} \\
  \midrule
  paper config                     & \textbf{5.2643} & \textbf{0.0000} & \textbf{0.7803} & 0.6118 & \textbf{5.08} \\
  w/o eta warmup (no warmup)       & 5.0345 & -0.2298 & 0.7803 & 0.6118 & 9.47 \\
  w/o decisive weighting ($w{=}1$) & 4.9029 & -0.3614 & 0.6789 & \textbf{0.6604} & 7.83 \\
  \bottomrule
  \end{tabular}%
  }
  \vspace{-8pt}
  \end{table}

\section{Conclusion}

We presented DynT2I-Eval, a dynamic evaluation framework for T2I models that replaces fixed prompt-set benchmarking with continual prompt generation, pairwise comparison, and online ranking updates. Using structured prompt spaces derived from long-form descriptions, it produces fresh samples for evaluating text alignment, perceptual quality, and aesthetic quality, while maintaining an updatable leaderboard through online scheduling and weighted Bayesian ranking. This dynamic protocol reduces prompt-set overfitting and provides a more sustainable evaluation setting for rapidly evolving T2I models, though results still depend on the evaluator backend in each dimension.

\paragraph{Limitations and Future Work.}
DynT2I-Eval improves robustness to benchmark-specific overfitting, but it does not eliminate all evaluation bias, particularly bias from imperfect automatic evaluators. Our goal is not to replace human judgment, but to provide a more robust protocol under a fixed evaluator; human evaluation therefore remains an important complement, especially for nuanced aspects of image quality and alignment. Because the framework is largely orthogonal to evaluator choice, stronger multimodal judges, better human-calibrated metrics, or evaluator ensembles could be introduced in future work to further improve reliability.
\bibliographystyle{plain} % 或者使用 neurips_2024 等模板指定的格式
\bibliography{ref}       % 注意这里的 refs 对应你的文件名 refs.bib
%%%%%%%%%%%%%%%%%%%%%%%%%%%%%%%%%%%%%%%%%%%%%%%%%%%%%%%%%%%%

%%%%%%%%%%%%%%%%%%%%%%%%%%%%%%%%%%%%%%%%%%%%%%%%%%%%%%%%%%%%

\newpage
\clearpage
\appendix

\section{Broader Impacts}
Our proposed dynamic evaluation framework offers a positive societal impact by mitigating benchmark contamination, ensuring that text-to-image models are evaluated on their true open-world generalization capabilities rather than their ability to overfit static prompt sets.

However, we also acknowledge potential negative impacts. Relying heavily on automated LLM-based and multimodal evaluators may inadvertently introduce or amplify the inherent biases present in those evaluator models (\textit{e.g.}, cultural, demographic, or aesthetic biases learned during the evaluator's pre-training). Furthermore, as our dynamic prompt generator creates open-ended scenarios, there is a theoretical risk of generating prompts that probe biased concepts. We mitigate this by using curated, benign long-form descriptions (\textit{e.g.}, DOCCI) as our semantic source, but future deployments of automated dynamic evaluators must remain vigilant about inheriting biases from the judge models.

\section{Human Evaluation Details and Ethical Considerations}
\subsection{Annotator Compensation and Interface}
To validate the reliability of our automatic evaluators, we conducted a small-scale human evaluation. The human annotators were compensated at a rate of approximately \$10 USD per hour, which is significantly above the local minimum wage, ensuring fair labor practices in accordance with the NeurIPS Code of Ethics.

During the evaluation, annotators were presented with a user interface displaying the text prompt alongside the generated Image A and Image B. They were instructed to select “Image A”, “Image B”, or “Tie” based strictly on the dimension being evaluated (text alignment, perceptual quality, or aesthetic quality). For the alignment task, annotators were explicitly provided with the atomic checklist and instructed to ignore stylistic preferences, focusing solely on whether the visual constraints were satisfied.

\subsection{Ethical Considerations and IRB Exemption}
This human validation involved solely the subjective evaluation of AI-generated images based on benign, safe text prompts. No personally identifiable information (PII) or sensitive demographic data was collected from the participants. Furthermore, the generated content was heavily constrained by our semantic deconstruction pipeline, preventing the generation of unsafe, explicit, or hazardous material. Given that the task poses no physical, psychological, or privacy risks to the annotators, this study is considered minimal risk and is typically exempt from formal Institutional Review Board (IRB) approval under standard institutional guidelines.

\section{Computational Resources}
All experiments, including text-to-image model inference, dynamic prompt generation via LLMs, and automatic multimodal evaluations, were conducted using NVIDIA RTX A6000 (48GB) GPUs.

Because our framework involves continuous, dynamic generation and our study encompasses an extensive set of experiments—including the main dynamic leaderboard construction across three dimensions for 12 models, large-scale transferability analyses, and comprehensive Monte Carlo ranking simulations—we did not track the precise aggregate GPU hours. However, the primary computational bottleneck lies in the image generation process and the inference of the multimodal evaluators (\textit{e.g.}, Qwen3.5 and VisualQuality-R1). Once the image pairs are generated and scored, our proposed Bayesian ranking updates and UCB-based scheduling are computationally lightweight and execute in seconds on a standard CPU.

\section{Additional Details of Dynamic Prompt Generation}
\label{app:prompt}

\subsection{Prompt Generation Pipeline}

DynT2I-Eval uses a dynamic prompt engine designed for continual benchmark expansion rather than one-shot prompt-list construction. The goal is not to randomly concatenate visual keywords, but to build a controlled and extensible prompt stream with diverse semantic coverage, adjustable difficulty, and strong discriminative value for model evaluation. The generator is shared across benchmark rounds and can be continuously expanded without changing the underlying construction policy.

The pipeline consists of five stages: (1) semantic decomposition of long-form image descriptions, (2) candidate-pool rebucketing and cleaning, (3) construction of task-specific sampling spaces, (4) online prompt instantiation for alignment and quality evaluation, and (5) automatic output analysis for quality control. This design turns long-form descriptions into a structured visual semantic space and reorganizes that space by samplability, judgeability, and task relevance before rewriting sampled configurations into natural-language prompts.

\subsection{Semantic Decomposition from Long-form Descriptions}

Our prompt source is a collection of long-form image descriptions. Each description is decomposed into six visual dimensions: \textit{subject}, \textit{cognitive logic constraints}, \textit{environment}, \textit{lighting and atmosphere}, \textit{camera and composition}, and \textit{medium and format}. The purpose of this decomposition is not to preserve the original description verbatim, but to convert dense captions into structured intermediate representations that can be sampled, recombined, and filtered under explicit control.

This representation serves three purposes. First, it exposes direct control interfaces for dynamic sampling. Second, it disentangles factors such as object identity, logical constraints, scene context, style, and viewpoint. Third, it allows different evaluation tasks to be constructed from the same semantic source while maintaining different prompt distributions.

\subsection{Candidate Re-bucketing and Structural Cleaning}

The decomposed candidates are not used directly. Instead, we reorganize them into benchmark-oriented buckets in order to separate expressions that merely sound descriptive from those suitable for dynamic and judgeable evaluation.

For the \textbf{subject} dimension, candidates are divided into several categories, including strong subjects, text-centric subjects, weak scene subjects, mixed subjects, uncertain subjects, and dependent fragments that require additional anchors. In formal prompt construction, we primarily sample from the strong-subject pool. A strong subject is intended to be visually salient, semantically self-contained, and easy to verify in the generated image. Weak scene subjects and dependent fragments are not discarded, but are kept in restricted intermediate pools rather than being allowed to dominate the main prompt distribution.

For the \textbf{logic} dimension, we regroup constraints into functional buckets including spatial relations, counting, short text rendering, long text rendering, attribute binding, state or action understanding, and other visual constraints. This logic rebucketing is the main control layer of difficulty. In particular, attribute binding and state or action constraints are better suited to easier alignment cases, counting and short text rendering provide stable medium-to-hard discrimination, spatial relations remain challenging for stronger models, and long text rendering is difficult enough to merit a dedicated mode.

\subsection{Task-specific Prompt Spaces}

We explicitly construct two different prompt spaces rather than mixing all evaluation goals into a single prompt distribution.

\paragraph{Alignment space.}
The alignment prompt space is designed for semantic faithfulness evaluation. Its objective is to produce prompts that are constraint-rich, visually judgeable, and discriminative, while avoiding unnecessary semantic noise. Each alignment prompt is built from three components: one subject tag, one logic-constraint tag, and one supporting tag. The subject is typically drawn from the strong-subject pool, the logic tag is sampled from a difficulty-controlled logic bucket, and the support tag is selected from one of four dimensions: \textit{environment}, \textit{lighting and atmosphere}, \textit{camera and composition}, or \textit{medium and format}.

\paragraph{Quality space.}
The quality prompt space is designed for perceptual-quality and aesthetic-quality evaluation, where overly heavy semantic load can interfere with the target judgment. Its objective is therefore to produce prompts that remain natural and visually rich while placing less weight on explicit logical constraints. Each quality prompt is typically built from one strong subject and two supporting dimensions, mainly from \textit{environment}, \textit{lighting and atmosphere}, and \textit{camera and composition}. A soft logic constraint may be added with low probability. The \textit{medium and format} dimension is used only occasionally, to avoid excessive stylization biasing quality evaluation.

\subsection{Difficulty-aware Online Sampling}

For alignment evaluation, we organize the prompt space into four difficulty modes: \textit{easy}, \textit{medium}, \textit{hard}, and \textit{hard long-text}. Difficulty is controlled primarily through the type of logic constraint rather than superficial wording variation.

The \textit{easy} mode mainly uses attribute binding and state or action constraints, and serves as a basic capability-check and judge-calibration layer. The \textit{medium} mode adds counting and short text rendering, creating prompts with a meaningful but still moderate semantic burden. The \textit{hard} mode emphasizes spatial relations, counting, and short text rendering, which better separate stronger text-to-image models. The \textit{hard long-text} mode is reserved for long text rendering only, targeting completeness, spelling stability, and readability of longer strings or multiword text.

The current alignment main set is generated through online dynamic sampling rather than blockwise generation. Before each new sample is created, the system first draws one of the four difficulty modes independently, using a uniform $25\%/25\%/25\%/25\%$ mixture. Conditioned on the sampled difficulty, it then selects a subject, a logic tag, and a supporting tag from the corresponding candidate pools, and finally rewrites the sampled semantic configuration into natural language using an LLM. Because difficulty is sampled independently for each item, the resulting prompt stream is naturally interleaved rather than divided into contiguous easy or hard segments. This avoids file-order-induced difficulty bias and supports continual expansion under a fixed sampling policy.

\subsection{Checklist Construction}

Alignment prompts are generated together with atomic checklists. Each checklist typically contains 4--6 items. The design principle is that every item should correspond to a visible fact, remain as atomic as possible, and cover a key part of the prompt semantics. We do not maximize checklist length; instead, we aim to cover the essential verifiable constraints with minimal redundancy. In practice, four items are usually sufficient for simpler prompts, while harder prompts---especially those involving spatial relations or long text rendering---often require a fifth or sixth item for adequate coverage.

\subsection{Stable Generation Strategy for Quality Prompts}

Compared with alignment prompt generation, the quality generator uses a more conservative strategy to improve output stability. Specifically, it targets shorter prompts, uses simpler and more direct instructions, disables logic constraints by default, keeps the logic-injection probability low, and retries generation when the output appears abnormal. During retry, the sampling space is further contracted if necessary. The goal is not to make quality prompts easier in a trivial sense, but to keep them natural, stable, and suitable for perceptual or aesthetic judgment without drifting back into alignment-heavy evaluation.

\subsection{Output Format and Engineering Support}

The generator writes prompts into append-only JSONL files. For alignment, the main output file stores, for each sample, its identifier, difficulty label, logic pool, source tags, rewritten prompt, and checklist. The quality generator uses the same append-only organization without overwriting existing results. The system also supports resume mode: when generation is interrupted, it detects the current maximum identifier in the output file and continues from the next sample. In addition, progress is flushed after each sample so that long-running generation remains observable.

\subsection{Automatic Output Analysis}

To diagnose generation quality, we use an automatic analysis module that reports statistics such as non-empty prompt rate, malformed-output rate, average prompt length, difficulty or logic-pool distribution, and template-pollution hits. The template-pollution check is used to detect low-value repeated phrasing patterns that would indicate degeneration toward templated prompt writing. This analysis converts generation quality from an informal impression into a set of observable indicators and serves as a safeguard for continual benchmark expansion.

\subsection{Prompt-space Size}

Under the current candidate pools and validity constraints, the structured configuration space is already very large before LLM rewriting. In the current implementation, the approximate configuration counts are on the order of $3.41\times 10^{12}$ for easy alignment, $9.33\times 10^{12}$ for medium alignment, $2.17\times 10^{13}$ for mixed hard alignment, $2.27\times 10^{11}$ for hard long-text alignment, and $3.72\times 10^{12}$ for the main quality space without logic expansion. If optional soft logic in the quality space is fully expanded, the upper bound becomes substantially larger. Although this space is not mathematically infinite, it is already large enough in practice to make prompt-level memorization or prompt-set-specific optimization much less effective than under a frozen benchmark list.

\subsection{Prompt-source Statistics}

The current prompt engine is built from 14{,}847 long-form source descriptions. After six-dimensional semantic decomposition and benchmark-oriented rebucketing, the resulting candidate pools become substantially larger than the number of source descriptions, because a single description may yield multiple usable candidates in different dimensions or multiple logic constraints within the same dimension.

In the current implementation, the alignment space contains 9{,}155 strong subject candidates, together with 7{,}486 attribute-binding candidates, 1{,}857 state/action candidates, 12{,}094 counting candidates, 5{,}948 short-text-rendering candidates, 40{,}627 spatial-relation candidates, and 621 long-text-rendering candidates. These counts are reported at the candidate level after decomposition and cleaning, rather than at the source-description level. As a result, they should be interpreted as the size of the structured sampling inventory available to the generator, not as the number of original captions.

Two observations are worth noting. First, the spatial-relation pool is substantially larger than the other logic pools, reflecting the fact that long-form descriptions often contain multiple extractable relational statements. Second, the long-text-rendering pool is much smaller than the short-text and spatial pools, which is consistent with the relative rarity of long readable text spans in naturally occurring image descriptions. This asymmetry is one reason why we isolate long-text rendering into a dedicated high-difficulty mode rather than treating it as an ordinary variant of generic logic constraints.

\subsection{Examples of Generated Alignment Prompts}

Below we show representative examples of dynamically generated alignment prompts. These examples illustrate that the final prompts are written in natural language rather than exposed as templated tag concatenations, and that the same structured pipeline can produce prompt types with different semantic burdens while remaining paired with atomic checklists.

\paragraph{Example 1: Easy / Attribute binding.}
\textbf{Prompt:}
A light gray brick wall building stands on a cement ground featuring dark and light gray shades. The structure includes large windows that run horizontally across its facade.

\textbf{Checklist:}
(1) Is the building wall constructed from light gray bricks?
(2) Do the windows run horizontally across the facade?
(3) Is the ground surface made of cement with mixed dark and light gray shades?
(4) Are the spatial relationships between the building and ground physically plausible?

\paragraph{Example 2: Medium / Attribute binding.}
\textbf{Prompt:}
A close-up view of a red frog's toes resting against a plain wall in the background. The frog's skin shows slight burning marks with a light tan coloration on the toes.

\textbf{Checklist:}
(1) Are the toes red?
(2) Is there visible burning on the skin?
(3) Is the burnt area light tan?
(4) Is a wall visible in the background?

\paragraph{Example 3: Hard / Short text rendering.}
\textbf{Prompt:}
A profile perspective view of a rounded flower bed edge. To the left, a recycling bin stands with a blue circular lid. The bin features the word ``recycle'' printed clearly in white letters on its side.

\textbf{Checklist:}
(1) Is the camera angle a clear profile view of the flower bed?
(2) Is the recycling bin positioned to the left of the flower bed?
(3) Does the bin have a blue circular lid?
(4) Is the word ``recycle'' visible on the side of the bin?
(5) Are the materials of the bin and plants depicted realistically?

\paragraph{Example 4: Hard long-text / Long text rendering.}
\textbf{Prompt:}
A close-up view of a statue's hand holding a small bird. To the left, wooden shelves are partially cut off by the frame edge. On the statue's chest, a label clearly reads ``9. american, ornate pair of demitasse cups and saucers''.

\textbf{Checklist:}
(1) Is the bird physically held in the statue's hand?
(2) Are the shelves on the left side partially cut off by the image frame?
(3) Does the visible text explicitly read ``9. american, ornate pair of demitasse cups and saucers''?
(4) Is the text rendered legibly on the statue's surface?
(5) Do the materials appear physically plausible?

\section{Additional Details of Evaluators}
\label{app:evaluators}

\subsection{Alignment Evaluation}

For text-image alignment, we use an LLM-as-a-judge protocol in a pairwise setting. Given a text prompt, its accompanying checklist, and two generated images, the judge is asked to determine which image better satisfies the prompt constraints. The evaluation is explicitly restricted to semantic faithfulness rather than overall visual appeal.

The judge is instructed to use the text prompt as the primary task definition and the checklist as grounding anchors for key visible constraints. It is further instructed not to reward beauty, mood, style, sharpness, or realism unless such factors directly affect whether the prompt is visibly satisfied. If both images satisfy the prompt to a similar degree, or if the evidence is insufficient, the judge returns \textit{Tie}.

The exact judge template is shown below.

\begin{quote}
\small
\texttt{You are an impartial judge for text-image alignment.}\\
\texttt{Your task is to compare Image A and Image B generated from the same prompt.}\\
\texttt{}\\
\texttt{Evaluation policy:}\\
\texttt{1. Use the Text Prompt as the primary task definition.}\\
\texttt{2. Use the Checklist as grounding anchors for the key visible constraints.}\\
\texttt{3. Judge only text-image alignment. Do not reward general beauty, mood, style, sharpness, or realism unless they directly affect whether the prompt constraints are visibly satisfied.}\\
\texttt{4. Prefer the image that better satisfies the prompt's core visible constraints, even if the other image looks more attractive overall.}\\
\texttt{5. If both images satisfy the prompt to a similar degree, or the evidence is insufficient, return Tie.}\\
\texttt{}\\
\texttt{[Text Prompt]}\\
\texttt{\{prompt\_text\}}\\
\texttt{}\\
\texttt{[Checklist]}\\
\texttt{\{checklist\_str\}}\\
\texttt{}\\
\texttt{Return STRICT JSON only:}\\
\texttt{\{}\\
\texttt{\ \ \ \ "analysis\_A": "one short sentence about Image A's alignment",}\\
\texttt{\ \ \ \ "analysis\_B": "one short sentence about Image B's alignment",}\\
\texttt{\ \ \ \ "winner": "A" or "B" or "Tie"}\\
\texttt{\}}
\end{quote}

\subsection{Perceptual Quality Evaluation}
\label{sec:perceptual_quality_evaluation_appendix}

Perceptual quality is evaluated independently of alignment. For each image, we first obtain repeated single-image quality scores and then convert them into a pairwise outcome through a variance-aware comparison rule.

Specifically, for each image we query the evaluator with $K=4$ repeated samples. Let the returned scores be $s_1, s_2, s_3, s_4$. We compute the mean score
\begin{equation}
\mu = \frac{1}{4}\sum_{i=1}^{4} s_i,
\label{eq:app_pq_mean}
\end{equation}
and the sample variance
\begin{equation}
\mathrm{var} = \frac{1}{4-1}\sum_{i=1}^{4}(s_i-\mu)^2.
\label{eq:app_pq_variance}
\end{equation}

Given two images $A$ and $B$, with corresponding statistics $(\mu_A, \mathrm{var}_A)$ and $(\mu_B, \mathrm{var}_B)$, we compute the pairwise win probability as
\begin{equation}
P(A \succ B)=\Phi\left(
\frac{\mu_A-\mu_B}
{\sqrt{\max(0,\mathrm{var}_A)+\max(0,\mathrm{var}_B)+\epsilon}}
\right),
\label{eq:app_pq_pairwise_prob}
\end{equation}
where $\Phi(\cdot)$ denotes the standard normal cumulative distribution function and $\epsilon=10^{-5}$ is a small constant for numerical stability.

We then convert this probability into a ternary decision:
\begin{equation}
\mathrm{winner} =
\begin{cases}
A, & \text{if } P(A \succ B) > 0.58,\\
B, & \text{if } P(A \succ B) < 0.42,\\
\mathrm{Tie}, & \text{otherwise.}
\end{cases}
\label{eq:app_pq_ternary_decision}
\end{equation}

This rule is intentionally conservative: larger uncertainty pushes the pairwise probability toward $0.5$, and small score differences are not over-interpreted as decisive wins.

\subsection{Aesthetic Quality Evaluation}

Aesthetic quality is evaluated as a separate dimension. In the current implementation, the evaluator produces scalar aesthetic scores, which are converted into a ternary pairwise outcome using a fixed empirical tie threshold. Specifically, if the absolute score difference between two images is smaller than 3.5 on the underlying 100-point scale, the result is treated as \textit{Tie}; otherwise, the higher-scoring image is preferred.

\subsection{Order-swapped Pairwise Aggregation}

For pairwise judgments, we evaluate each image pair twice by swapping the display order of the two candidates. This is done to reduce order sensitivity in multimodal judgment models. For an image pair $(A,B)$, we obtain one decision from the presentation order $(A,B)$ and another from $(B,A)$, and then map both outputs back to the same canonical label space $\{A, B, \mathrm{Tie}\}$.

The final result is aggregated as follows. If the two judgments agree, we keep that result. If they disagree and one judgment is \textit{Tie}, we return the non-tie result. If they disagree and neither judgment is \textit{Tie}, we return \textit{Tie}. This conservative rule reduces order bias without forcing unstable wins.

\section{Scheduling and Ranking Algorithms}
\label{app:ranking}

This appendix provides the full mathematical specification of the online scheduling, micro-batch aggregation, and Bayesian rating update procedures used in DynT2I-Eval.

\subsection{Rating State and Hyperparameters}
\label{app:rating_state}

Each model $i$ is associated with a Gaussian-style rating state parameterized by a mean $\mu_i$ and a standard deviation $\sigma_i$. At initialization, all models share the same prior:
\begin{equation}
\mu_i = \mu_0, \qquad \sigma_i = \sigma_0,
\label{eq:init_rating}
\end{equation}
where in our implementation
\begin{equation}
\mu_0 = 1000, \qquad \sigma_0 = 300.
\label{eq:init_values}
\end{equation}

The rating system further uses the following hyperparameters:
\begin{equation}
\beta = 70,\quad
\eta = 3,\quad
\alpha = 190,\quad
\sigma_{\mathrm{conv}} = 20,
\label{eq:hyper_1}
\end{equation}
\begin{equation}
p_{\mathrm{low}} = 0.42,\quad
p_{\mathrm{high}} = 0.58,\quad
\rho_{\mathrm{tie}} = 0.98,\quad
\lambda_{\mathrm{tie}} = 0.05,
\label{eq:hyper_2}
\end{equation}
\begin{equation}
s_w = 2.0,\quad
w_{\max} = 2.0,\quad
\gamma_{\sigma} = 1.0,\quad
M_{\mathrm{warm}} = 20,\quad
r_{\min} = 0.20.
\label{eq:hyper_3}
\end{equation}

Here, $\beta$ controls the scale of pairwise performance noise in the Bayesian update, $\alpha$ controls exploration in opponent selection, and $\sigma_{\mathrm{conv}}$ defines the minimum uncertainty floor. The thresholds $p_{\mathrm{low}}$ and $p_{\mathrm{high}}$ determine whether a micro-batch is resolved as a decisive win or a tie. The parameters $\rho_{\mathrm{tie}}$ and $\lambda_{\mathrm{tie}}$ govern the tie update, while $s_w$ and $w_{\max}$ determine the strength of the batch-level update weight.

\subsection{Conservative Ranking Score}
\label{app:conservative_score}

Although the posterior state of a model is represented by $(\mu_i,\sigma_i)$, leaderboard ranking is based on a conservative score rather than on $\mu_i$ alone. Let $m_i$ denote the number of evaluated micro-batches involving model $i$. We define an effective conservativeness coefficient
\begin{equation}
\eta_{\mathrm{eff}}(m_i)=
\begin{cases}
\eta, & \text{if } M_{\mathrm{warm}} \le 0,\\[4pt]
\eta \cdot \max\!\left(r_{\min},\, \min\!\left(1,\frac{m_i}{M_{\mathrm{warm}}}\right)\right), & \text{otherwise.}
\end{cases}
\label{eq:eta_eff}
\end{equation}
The conservative score of model $i$ is then
\begin{equation}
S_i = \mu_i - \eta_{\mathrm{eff}}(m_i)\sigma_i.
\label{eq:conservative_score}
\end{equation}
Models are ranked in descending order of $S_i$.

\subsection{Online Match Scheduling}
\label{app:online_scheduling}

The benchmark proceeds in rounds. At each round, the scheduler selects one pair of models, samples a micro-batch of prompts, obtains prompt-level comparison outcomes, and then performs one batch-level rating update.

Let $\mathcal{M}_{\mathrm{act}}$ denote the set of active models and $\mathcal{M}_{\mathrm{conv}}$ the set of converged models. To ensure balanced participation, the scheduler first selects a pivot model as the active model with the fewest evaluated micro-batches:
\begin{equation}
i^\star = \arg\min_{i\in \mathcal{M}_{\mathrm{act}}} c_i.
\label{eq:pivot_selection}
\end{equation}

If multiple active models remain, an opponent is selected by a UCB-style criterion. For each candidate opponent $j\neq i^\star$, we define
\begin{equation}
\mathrm{exploit}(i^\star,j) = -|\mu_{i^\star} - \mu_j|,
\label{eq:exploit_term}
\end{equation}
\begin{equation}
\mathrm{explore}(i^\star,j) = \alpha \sqrt{\frac{\log(\max(1,r))}{\max(1,n_{i^\star j})}},
\label{eq:explore_term}
\end{equation}
where $r$ is the current round index and $n_{i^\star j}$ is the number of previous encounters between models $i^\star$ and $j$.

The scheduler then constructs the lexicographically compared tuple
\begin{equation}
\Pi(i^\star,j)=\left(
u_{i^\star j},
b_j,
\xi_{i^\star j}
\right),
\label{eq:selection_tuple}
\end{equation}
with
\begin{equation}
u_{i^\star j}=\mathrm{exploit}(i^\star,j)+\mathrm{explore}(i^\star,j),
\label{eq:ucb_main_score}
\end{equation}
\begin{equation}
b_j = \sigma_j^{\gamma_{\sigma}},
\label{eq:sigma_bias}
\end{equation}
and
\begin{equation}
\xi_{i^\star j} \sim \mathrm{Unif}(0,1).
\label{eq:jitter}
\end{equation}
The selected opponent is
\begin{equation}
j^\star = \arg\max_{j\in \mathcal{M}_{\mathrm{act}}\setminus\{i^\star\}} \Pi(i^\star,j),
\label{eq:opponent_selection}
\end{equation}
where the comparison is lexicographic: the scheduler first maximizes $u_{i^\star j}$, then favors larger uncertainty bias $b_j$, and finally uses the random jitter $\xi_{i^\star j}$ to break exact ties.

If only one active model remains, the scheduler assigns a sparring opponent from the converged set whose posterior mean is closest:
\begin{equation}
j^\star = \arg\min_{j\in \mathcal{M}_{\mathrm{conv}}} |\mu_j-\mu_{i^\star}|.
\label{eq:sparring_partner}
\end{equation}

After selecting the pair $(i^\star,j^\star)$, the system samples a micro-batch of $B=15$ previously unused prompts.

\subsection{Micro-Batch Aggregation}
\label{app:micro_batch}

A model pair is compared on a micro-batch of $B=15$ prompts. Let the prompt-level outcomes be $\{y_b\}_{b=1}^B$, with each
\[
y_b \in \{A,B,\mathrm{Tie}\}.
\]
We summarize the micro-batch by the counts
\begin{equation}
w_A = \sum_{b=1}^{B} \mathbb{I}[y_b = A], \qquad
w_B = \sum_{b=1}^{B} \mathbb{I}[y_b = B], \qquad
\tau = \sum_{b=1}^{B} \mathbb{I}[y_b = \mathrm{Tie}],
\label{eq:micro_batch_counts}
\end{equation}
and
\begin{equation}
N = w_A + w_B + \tau.
\label{eq:total_valid}
\end{equation}

If $N=0$, the batch is treated as invalid and no meaningful comparison result is produced. Otherwise, we define the empirical batch-level score for model $A$ as
\begin{equation}
p_A = \frac{w_A + 0.5\,\tau}{N}.
\label{eq:batch_score_pa}
\end{equation}
This quantity treats each tie as half a win for each side.

The batch-level macro outcome is then resolved by a two-threshold rule:
\begin{equation}
\mathrm{macro}(A,B) =
\begin{cases}
A, & \text{if } p_A > p_{\mathrm{high}},\\[4pt]
B, & \text{if } p_A < p_{\mathrm{low}},\\[4pt]
\mathrm{Tie}, & \text{otherwise.}
\end{cases}
\label{eq:macro_outcome}
\end{equation}

\subsection{Batch Weighting}
\label{app:batch_weight}

The strength of the posterior update is modulated by a batch weight derived from the decisiveness of the micro-batch. Define
\begin{equation}
p_w =
\begin{cases}
p_A, & \text{if } \mathrm{macro}(A,B)=A,\\[4pt]
1-p_A, & \text{if } \mathrm{macro}(A,B)=B,\\[4pt]
0.5, & \text{if } \mathrm{macro}(A,B)=\mathrm{Tie}.
\end{cases}
\label{eq:pw}
\end{equation}
The batch weight is then
\begin{equation}
\omega = \min\!\left(\omega_{\max},\, 1 + \lambda_w(p_w - 0.5)\right).
\label{eq:batch_weight_eq}
\end{equation}
Under the default configuration, $1 \le \omega \le 2$.

\subsection{Bayesian Rating Update for Decisive Batches}
\label{app:bayesian_update_winloss}

If the macro outcome is decisive, the batch is treated as a weighted pairwise win/loss observation. Suppose model $w$ is the winner and model $l$ is the loser. Let their current states be $(\mu_w,\sigma_w)$ and $(\mu_l,\sigma_l)$.

We first compute
\begin{equation}
c = \sqrt{\sigma_w^2 + \sigma_l^2 + 2\beta^2},
\label{eq:c12}
\end{equation}
and
\begin{equation}
t = \mathrm{clip}\!\left(\frac{\mu_w-\mu_l}{c},\, -5,\, 5\right).
\label{eq:t_value}
\end{equation}
Let $\phi(\cdot)$ and $\Phi(\cdot)$ denote the standard Gaussian PDF and CDF, respectively. Define
\begin{equation}
V(t) = \frac{\phi(t)}{\max(\Phi(t),10^{-12})},
\qquad
W(t) = V(t)\bigl(V(t)+t\bigr).
\label{eq:vw}
\end{equation}

The posterior means are updated as
\begin{equation}
\mu_w' = \mu_w + \frac{\sigma_w^2}{c} V(t)\, \omega,
\qquad
\mu_l' = \mu_l - \frac{\sigma_l^2}{c} V(t)\, \omega.
\label{eq:mu_update}
\end{equation}

To modulate uncertainty shrinkage according to batch decisiveness, we define
\begin{equation}
\Gamma(\omega) = \min\!\left(1.2,\, 1 + 0.5(\omega-1)\right).
\label{eq:gamma_weight}
\end{equation}
The posterior variances are then updated by
\begin{equation}
\sigma_w'^2 =
\max\!\left(
\sigma_{\mathrm{conv}}^2,\,
\sigma_w^2 \left[1-\frac{\sigma_w^2}{c^2}W(t)\Gamma(\omega)\right]
\right),
\label{eq:sigma_w_update}
\end{equation}
\begin{equation}
\sigma_l'^2 =
\max\!\left(
\sigma_{\mathrm{conv}}^2,\,
\sigma_l^2 \left[1-\frac{\sigma_l^2}{c^2}W(t)\Gamma(\omega)\right]
\right).
\label{eq:sigma_l_update}
\end{equation}
Finally,
\begin{equation}
\sigma_w' = \sqrt{\sigma_w'^2},
\qquad
\sigma_l' = \sqrt{\sigma_l'^2}.
\label{eq:sigma_final}
\end{equation}

This is a weighted TrueSkill-style pairwise Bayesian update with an explicit uncertainty floor at $\sigma_{\mathrm{conv}}$.

\subsection{Tie Update}
\label{app:tie_update}

If the macro outcome of a micro-batch is a tie, the system applies a symmetric update that slightly pulls the two means toward one another while reducing uncertainty.

Let the midpoint of the two means be
\begin{equation}
m = \frac{\mu_A+\mu_B}{2}.
\label{eq:tie_midpoint}
\end{equation}
The updated means are
\begin{equation}
\mu_A' = (1-\rho)\mu_A + \rho m,
\qquad
\mu_B' = (1-\rho)\mu_B + \rho m.
\label{eq:tie_mu_update}
\end{equation}
The posterior standard deviations are updated by
\begin{equation}
\sigma_A' =
\sqrt{
\max\!\left(
\sigma_{\mathrm{conv}}^2,\,
\rho_{\mathrm{tie}}\sigma_A^2
\right)
},
\qquad
\sigma_B' =
\sqrt{
\max\!\left(
\sigma_{\mathrm{conv}}^2,\,
\rho_{\mathrm{tie}}\sigma_B^2
\right)
}.
\label{eq:tie_sigma_update}
\end{equation}
\subsection{Statistics Recording versus Posterior Updating}
\label{app:stats_vs_update}

For each micro-batch, the system records the exact prompt-level counts $(W_A,W_B,T)$ as descriptive statistics. These counts are accumulated into the models' total numbers of wins, losses, ties, and evaluated prompts. However, the posterior state $(\mu,\sigma)$ is updated only once per micro-batch, using the aggregated macro outcome $\hat{y}$ in Eq.~\eqref{eq:macro_outcome} and the batch weight $w$ in Eq.~\eqref{eq:batch_weight_eq}.

This design preserves prompt-level evidence for later analysis while avoiding overly aggressive posterior drift that would arise if the 15 prompts in a single micro-batch were treated as fully independent Bayesian match outcomes.

\subsection{One-Round Procedure}
\label{app:one_round}

Algorithm~\ref{alg:online_benchmark_round} summarizes one round of the benchmark.

\begin{algorithm}[t]
\caption{One round of online benchmark scheduling and rating update}
\label{alg:online_benchmark_round}
\begin{algorithmic}[1]
\Require Active set $\mathcal{M}_{\mathrm{act}}$, converged set $\mathcal{M}_{\mathrm{conv}}$, rating states $\{(\mu_i,\sigma_i)\}_i$, prompt pool $\mathcal{P}$
\State Select pivot model:
\[
i^\star \gets \arg\min_{i\in\mathcal{M}_{\mathrm{act}}} m_i
\]
\If{$|\mathcal{M}_{\mathrm{act}}| > 1$}
    \State Select opponent:
    \[
    j^\star \gets \arg\max_{j\in\mathcal{M}_{\mathrm{act}}\setminus\{i^\star\}} U(i^\star,j)
    \]
\Else
    \State Select sparring partner:
    \[
    j^\star \gets \arg\min_{j\in\mathcal{M}_{\mathrm{conv}}} |\mu_j-\mu_{i^\star}|
    \]
\EndIf
\State Sample a micro-batch of $B=15$ unused prompts from $\mathcal{P}$
\State Generate one image per model for each prompt
\For{each prompt $b=1,\dots,B$}
    \State Obtain prompt-level comparison outcome $y_b \in \{A,B,\mathrm{Tie}\}$
\EndFor
\State Aggregate outcomes into $(W_A,W_B,T)$ using Eq.~\eqref{eq:micro_batch_counts}
\State Compute $p_A$ via Eq.~\eqref{eq:batch_score_pa}
\State Resolve macro outcome $\hat{y}$ via Eq.~\eqref{eq:macro_outcome}
\State Compute batch weight $w$ via Eq.~\eqref{eq:batch_weight_eq}
\If{$\hat{y}\in\{A,B\}$}
    \State Apply decisive Bayesian update using Eqs.~\eqref{eq:c12}--\eqref{eq:sigma_final}
\Else
    \State Apply tie update using Eqs.~\eqref{eq:tie_midpoint}--\eqref{eq:tie_sigma_update}
\EndIf
\For{each model $i$}
    \State Recompute conservative score:
    \[
    S_i \gets \mu_i - \eta_{\mathrm{eff}}(m_i)\sigma_i
    \]
\EndFor
\State Update leaderboard by sorting models in descending order of $S_i$
\end{algorithmic}
\end{algorithm}

\paragraph{Implementation note.}
The pseudocode above describes the logical ranking procedure. In the actual system implementation, image generation and evaluator invocation are executed asynchronously: once both images for a prompt are fully written to disk and verified as valid image files, the prompt is submitted for evaluation in the background. This concurrency mechanism improves system throughput but does not alter the mathematical definition of the ranking algorithm.

\section{Additional Eval Statistics}
\subsection{Distribution of Alignment Prompts}
\label{app:prompt_distribution}

To characterize the composition of the dynamic alignment benchmark, we summarize the distribution of prompts in the main alignment pool. In total, the pool contains 9,033 prompts. The difficulty levels are intentionally balanced, with 2,302 easy prompts (25.5\%), 2,290 medium prompts (25.4\%), 2,225 hard prompts (24.6\%), and 2,216 hard long-text prompts (24.5\%). This near-uniform split ensures that the benchmark does not over-emphasize any single difficulty regime.

We further examine the distribution over logic categories. The prompt pool covers six logic families: attribute binding (21.7\%), counting (16.7\%), spatial reasoning (8.3\%), state/action understanding (12.5\%), short text rendering (16.3\%), and long text rendering (24.5\%). As expected, long-text prompts occupy a relatively larger portion because they constitute an entire dedicated difficulty bucket. More broadly, the distribution indicates that the pool spans both classical compositional alignment challenges, such as binding and counting, and text-intensive scenarios that remain difficult for current text-to-image models.

In addition, the auxiliary support categories are also balanced. Camera and composition prompts account for 24.6\% of the pool, environment for 25.2\%, lighting and atmosphere for 25.1\%, and medium/format for 25.0\%. This balanced design prevents the benchmark from repeatedly relying on a narrow stylistic scaffold and encourages broader coverage of scene, style, and presentation conditions.

Finally, we analyze the checklist complexity associated with each prompt. Most prompts contain either four checklist items (40.9\%) or five checklist items (56.1\%), while only a small fraction contain six items (3.0\%). The average checklist length is 4.62 items per prompt, indicating that most prompts require multi-aspect verification rather than a single binary decision. Overall, these statistics show that the prompt pool is diverse in semantic content and well balanced in difficulty, logic type, support context, and evaluation granularity.

\begin{table}[t]
\centering
\small
\caption{Distribution of prompts in the main alignment pool.}
\label{tab:alignment_prompt_distribution}
\begin{tabular}{llrr}
\toprule
Category & Subcategory & Count & Ratio \\
\midrule
\multirow{4}{*}{Difficulty}
& Easy & 2302 & 0.2548 \\
& Medium & 2290 & 0.2535 \\
& Hard & 2225 & 0.2463 \\
& Hard long-text & 2216 & 0.2453 \\
\midrule
\multirow{6}{*}{Logic family}
& Attribute binding & 1962 & 0.2172 \\
& Counting & 1512 & 0.1674 \\
& Spatial reasoning & 749 & 0.0829 \\
& State/action understanding & 1125 & 0.1245 \\
& Short text rendering & 1469 & 0.1626 \\
& Long text rendering & 2216 & 0.2453 \\
\midrule
\multirow{4}{*}{Support category}
& Camera composition & 2224 & 0.2462 \\
& Environment & 2276 & 0.2520 \\
& Lighting atmosphere & 2272 & 0.2515 \\
& Medium format & 2261 & 0.2503 \\
\midrule
\multirow{3}{*}{Checklist count}
& 4 items & 3690 & 0.4085 \\
& 5 items & 5071 & 0.5614 \\
& 6 items & 272 & 0.0301 \\
\bottomrule
\end{tabular}
\end{table}

\subsection{Prompt Length Statistics}
\label{app:prompt_length}

We next report basic length statistics for the prompt pools used in DynT2I-Eval. For the main alignment pool, prompts contain on average 56.83 words and 334.09 characters. The corresponding average checklist length is 4.62 items. These numbers reflect the fact that alignment prompts are intentionally descriptive and structured, often combining a core instruction with several compositional constraints that must be checked individually.

By comparison, prompts in the main quality pool are shorter, with an average length of 40.69 words and 237.05 characters. This difference is consistent with the roles of the two prompt pools. Alignment evaluation requires prompts with explicit, decomposable requirements so that the evaluator can verify faithfulness to multiple conditions. Quality evaluation, in contrast, places greater emphasis on perceptual or aesthetic judgment and therefore does not require the same level of compositional specification.

Taken together, these statistics highlight a deliberate separation between the two prompt regimes: alignment prompts are longer and more checklist-driven, whereas quality prompts are shorter and comparatively open-ended. This design helps the benchmark probe distinct dimensions of model behavior without conflating instruction-following difficulty with perceptual or aesthetic preference.

\begin{table}[t]
\centering
\small
\caption{Length statistics of the prompt pools.}
\label{tab:prompt_length_stats}
\begin{tabular}{lrrr}
\toprule
Prompt pool & Avg. words & Avg. characters & Avg. checklist items \\
\midrule
Alignment main & 56.83 & 334.09 & 4.62 \\
Quality main   & 40.69 & 237.05 & -- \\
\bottomrule
\end{tabular}
\end{table}

\section{Dynamic Leaderboard Stability and Late-Entry Positioning}

\begin{figure}[t]
    \centering
    \includegraphics[width=1\linewidth]{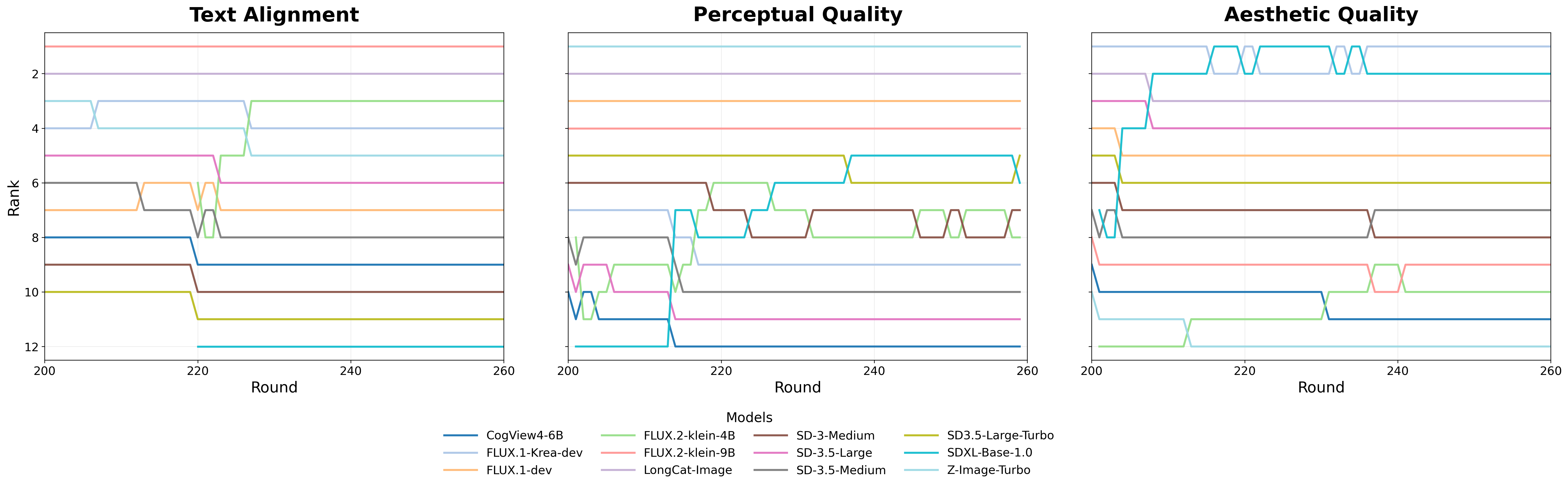}
    \caption{Model rank trajectories from rounds 200 to 260 across Text Alignment, Perceptual Quality, and Aesthetic Quality. By this stage, the leaderboard is already largely stable. The remaining fluctuations mainly occur between closely matched models, where adjacent ranks may still swap occasionally.}
    \label{fig:combined_rank}
\end{figure}

Figure~\ref{fig:combined_rank} visualizes model rank trajectories from rounds 200 to 260 across Text Alignment, Perceptual Quality, and Aesthetic Quality. The figure shows that by this stage the leaderboards are already largely stable, with the remaining variations mostly limited to occasional swaps between adjacent models of similar strength. This late-stage behavior is consistent with the quantitative convergence statistics in Table~\ref{tab:global_convergence_summary}.

To measure convergence more explicitly, we compare intermediate leaderboards with the final leaderboard in each dimension. The dynamic rankings stabilize quickly across all three dimensions. By round 100, the Spearman correlation with the final ranking reaches 1.000 for Text Alignment, 0.976 for Perceptual Quality, and 0.988 for Aesthetic Quality, while the Top-5 overlap is already 1.0 in all three cases. Text Alignment and Aesthetic Quality converge particularly quickly, both reaching a Spearman correlation of 0.964 by round 50. Perceptual Quality shows somewhat slower early-stage convergence (0.782 at round 50), but also approaches its final ordering rapidly afterward. At the round nearest to 250, the three dimensions are almost identical to their final rankings, with Spearman correlations of 0.993, 0.993, and 1.000 for Text Alignment, Perceptual Quality, and Aesthetic Quality, respectively. These results indicate that the leaderboard can already provide highly reliable relative ordering well before the full evaluation budget is exhausted.

We further examine whether the framework can efficiently accommodate newly added models by injecting two late-entry anchors, \textsc{SDXL-Base-1.0} and \textsc{FLUX.2-klein-4B}, after the leaderboard has largely formed. The injection round is 220 for Text Alignment and 201 for Perceptual Quality and Aesthetic Quality. Across the six late-entry trajectories, the injected models enter the final-rank $\pm 1$ neighborhood in 6.5 rounds on average, and reach a stricter 11-round stable $\pm 1$ neighborhood in 12.5 rounds on average. In several cases, placement is nearly immediate: \textsc{SDXL-Base-1.0} is assigned its final Text Alignment rank at the injection round itself, and \textsc{FLUX.2-klein-4B} enters the final-rank $\pm 1$ neighborhood immediately in Perceptual Quality. Overall, these results suggest that the dynamic leaderboard not only converges rapidly for the original model pool, but can also localize the rank region of a newly introduced model with modest additional comparison cost. A detailed summary of late-entry positioning metrics is provided in Table~\ref{tab:late_entry_summary}.

\begin{table*}[t]
\centering
\caption{Round-level convergence statistics of the dynamic leaderboard, measured by comparison with the final ranking of the same dimension. Rank correlation becomes high early in all three dimensions, and by round 100 the Top-5 set already matches the final leaderboard exactly for Text Alignment, Perceptual Quality, and Aesthetic Quality. By the round nearest to 250, the leaderboards are nearly identical to their final orderings.}
\label{tab:global_convergence_summary}
\small
\resizebox{\textwidth}{!}{%
\begin{tabular}{llccccc}
\toprule
Dimension & Round & \# Active Models & Spearman & Kendall $\tau$ & Top-3 Overlap & Top-5 Overlap \\
\midrule
Aesthetic Quality   & 50  & 10 & 0.964 & 0.867 & 1.000 & 1.000 \\
Aesthetic Quality   & 100 & 10 & 0.988 & 0.956 & 1.000 & 1.000 \\
Aesthetic Quality   & 150 & 10 & 0.988 & 0.956 & 1.000 & 1.000 \\
Aesthetic Quality   & 200 & 10 & 0.988 & 0.956 & 1.000 & 1.000 \\
Aesthetic Quality   & 250 & 12 & 1.000 & 1.000 & 1.000 & 1.000 \\
\midrule
Perceptual Quality  & 50  & 10 & 0.782 & 0.600 & 0.667 & 1.000 \\
Perceptual Quality  & 100 & 10 & 0.976 & 0.911 & 1.000 & 1.000 \\
Perceptual Quality  & 150 & 10 & 0.976 & 0.911 & 1.000 & 1.000 \\
Perceptual Quality  & 200 & 10 & 1.000 & 1.000 & 1.000 & 1.000 \\
Perceptual Quality  & 250 & 12 & 0.993 & 0.970 & 1.000 & 0.800 \\
\midrule
Text Alignment      & 50  & 10 & 0.964 & 0.911 & 1.000 & 1.000 \\
Text Alignment      & 100 & 10 & 1.000 & 1.000 & 1.000 & 1.000 \\
Text Alignment      & 150 & 10 & 0.988 & 0.956 & 0.667 & 1.000 \\
Text Alignment      & 200 & 10 & 0.988 & 0.956 & 1.000 & 1.000 \\
Text Alignment      & 250 & 12 & 0.993 & 0.970 & 1.000 & 1.000 \\
\bottomrule
\end{tabular}%
}
\end{table*}

\begin{table*}[t]
\centering
\caption{Late-entry rank localization statistics for two injected models. ``Rounds to First Hit'' denotes the number of rounds required for the model to enter the final-rank $\pm 1$ neighborhood after injection. ``Rounds to Stable Hit'' denotes the number of rounds required before the model stays within that neighborhood for the following 10 rounds. The results show that newly added models are typically placed near their final rank with modest additional comparison budget.}
\label{tab:late_entry_summary}
\small
\resizebox{\textwidth}{!}{%
\begin{tabular}{llcccccc}
\toprule
Dimension & Model & Inj. Round & Final Rank & First Hit & Stable Hit & Rounds to First Hit & Rounds to Stable Hit \\
\midrule
Aesthetic Quality  & FLUX.2-klein-4B & 201 & 10 & 213 & 213 & 12 & 12 \\
Aesthetic Quality  & SDXL-Base-1.0   & 201 & 2  & 208 & 208 & 7  & 7  \\
\midrule
Perceptual Quality & FLUX.2-klein-4B & 201 & 8  & 201 & 227 & 0  & 26 \\
Perceptual Quality & SDXL-Base-1.0   & 201 & 6  & 214 & 224 & 13 & 23 \\
\midrule
Text Alignment     & FLUX.2-klein-4B & 220 & 3  & 227 & 227 & 7  & 7  \\
Text Alignment     & SDXL-Base-1.0   & 220 & 12 & 220 & 220 & 0  & 0  \\
\bottomrule
\end{tabular}%
}
\end{table*}

\section{Extended Robustness Analysis Across Six Stress-Test Environments}
\label{sec:appendix_six_envs}

To complement the main ablation and benchmark results in the paper, we provide a more detailed analysis of method behavior under six simulated evaluation environments. These environments are designed to stress different sources of uncertainty in dynamic pairwise evaluation, including judge noise, extreme-case perturbations, and prompt-level heterogeneity. The goal of this analysis is not to introduce new algorithmic components, but to examine whether the observed advantages of our method remain consistent when the evaluation process becomes substantially noisier, more heterogeneous, or more adversarial.

Across all environments, we compare the same set of methods: \textsc{Elo}, \textsc{TrueSkill}, \textsc{TrueSkill2}, \textsc{Glicko-2}, \textsc{Rank Centrality}, \textsc{K-Sort Arena}, \textsc{Active Ranking (AR)}, \textsc{RUCB}, and \textsc{Ours}. Each environment is evaluated over 20 independent trials. Every trial consists of a dynamic phase with 3000 comparison rounds, during which new strong models are periodically injected, followed by a short static phase of 50 rounds without further injection to assess final ranking quality. We report the mean rank correlation at 500, 1500, and 3000 rounds (SRCC@500, SRCC@1500, SRCC@3000), together with metrics that quantify dynamic adaptation: the average discovery latency of newly injected strong models, the discovery success rate, and the rate at which the injected model is eventually ranked first.

\paragraph{Environment design.}
The six environments are summarized in Table~\ref{tab:appendix_env_definitions}. Intuitively, they can be grouped into three families. The first family modifies the noise characteristics of the evaluator (\texttt{reference} and \texttt{noisier\_judge}). The second family changes the frequency and severity of hard cases (\texttt{hard\_extreme} and \texttt{less\_extreme}). The third family introduces prompt-dependent variation and structured prompt bias (\texttt{prompt\_calibrated} and \texttt{prompt\_clustered\_stress}). This setup allows us to test not only standard robustness to stochastic noise, but also sensitivity to distributional heterogeneity that is more realistic in LLM- or generation-based evaluation pipelines.

\begin{table*}[t]
\centering
\caption{Definitions of the six stress-test environments used in the extended robustness analysis. Parameter changes are described relative to the default \texttt{reference} environment.}
\label{tab:appendix_env_definitions}
\small
\resizebox{\textwidth}{!}{%
\begin{tabular}{p{2.5cm}p{3.0cm}p{5.8cm}p{5.6cm}}
\toprule
Environment & Perturbation type & Key parameter changes & Interpretation \\
\midrule
\texttt{reference} 
& Baseline setting 
& Default environment configuration: $\beta=75$, $\sigma_{\text{std}}=50$, $\sigma_{\text{ext}}=100$, $\mu_{\text{penalty}}=-100$, $\sigma_{\text{penalty}}=30$, tie threshold $=30$, extreme probability $=0.5$, no prompt-level perturbation 
& Standard benchmark environment without additional prompt heterogeneity or structured stress \\
\midrule
\texttt{noisier\_judge} 
& Higher judgment noise 
& Increased evaluator uncertainty and tie tendency: $\beta=95$, tie threshold $=35$ 
& Simulates a noisier judge that is both less decisive and more likely to return ties \\
\midrule
\texttt{hard\_extreme} 
& Stronger extreme-event disturbance 
& More severe and more frequent extreme cases: $\sigma_{\text{ext}}=130$, $\mu_{\text{penalty}}=-120$, $\sigma_{\text{penalty}}=40$, extreme probability $=0.65$ 
& Stress test with harsher rare events, larger perturbations, and stronger degradation under adverse cases \\
\midrule
\texttt{less\_extreme} 
& Milder stochastic environment 
& Reduced normal and extreme noise: $\sigma_{\text{std}}=45$, $\sigma_{\text{ext}}=80$, $\mu_{\text{penalty}}=-80$, tie threshold $=25$, extreme probability $=0.35$ 
& Easier environment with fewer extreme events and a lower tendency toward ambiguous outcomes \\
\midrule
\texttt{prompt\_calibrated} 
& Prompt-level heterogeneity 
& Introduces prompt-specific variation in difficulty and tie sensitivity: prompt difficulty std $=35$, prompt tie-sensitivity std $=8$; no cluster-level bias 
& Different prompts induce different difficulty levels and decisiveness, but without structured prompt groups \\
\midrule
\texttt{prompt\_clustered\_}\newline\texttt{stress}
& Prompt-level heterogeneity + structured bias 
& Builds on \texttt{prompt\_calibrated} and further adds prompt clusters: prompt cluster count $=4$, prompt cluster scale $=45$ 
& A stronger stress-test setting in which prompts belong to clusters that systematically favor or penalize subsets of models \\
\bottomrule
\end{tabular}%
}
\end{table*}

\paragraph{Main observations.}
The detailed results are reported in Table~\ref{tab:appendix_six_env_results}. Several patterns are consistent across environments.

First, our method achieves the best \emph{early-stage} ranking quality in all six environments. In particular, \textsc{Ours} is the top method at both SRCC@500 and SRCC@1500 for every environment considered here. This pattern is important because, in dynamic evaluation settings, rankings are often consumed before the system has fully converged. Strong early-stage SRCC therefore indicates that the method can recover the global ordering much faster under a fixed comparison budget.

Second, \textsc{Ours} maintains highly stable dynamic discovery behavior across all environments. Its average discovery latency stays in a narrow range of roughly 5.7--6.1 rounds, and the discovery success rate remains at $100\%$ in all six settings. This consistency is notable because several baselines exhibit large changes in discovery performance when the environment becomes noisier or more heterogeneous. For example, \textsc{RUCB} remains competitive in some environments but experiences a clear slowdown under prompt-cluster stress, while \textsc{K-Sort Arena} consistently requires many more rounds to discover injected strong models despite its strong eventual ranking quality.

Third, the final-stage ranking metric SRCC@3000 reveals a more nuanced trade-off. \textsc{K-Sort Arena} attains the best SRCC@3000 in several environments, especially in \texttt{reference}, \texttt{noisier\_judge}, \texttt{hard\_extreme}, \texttt{less\_extreme}, and \texttt{prompt\_clustered\_stress}. However, this higher final precision comes at the cost of much slower dynamic adaptation, with discovery latency typically around 26--27 rounds. By contrast, \textsc{Ours} sacrifices a small amount of asymptotic ranking precision in some settings in exchange for dramatically faster identification of newly injected strong models. This is precisely the regime we care about in dynamic leaderboards, where responsiveness and ranking quality must be balanced rather than optimized in isolation.

Finally, the most difficult environment is clearly \texttt{prompt\_clustered\_stress}. In this setting, all methods degrade substantially, indicating that structured prompt-dependent bias is a challenging source of distribution shift for dynamic ranking algorithms. Even so, \textsc{Ours} still achieves the best SRCC@500 and SRCC@1500, while preserving perfect discovery success. This suggests that our method is particularly strong at maintaining useful early-stage rankings even when final asymptotic ordering becomes difficult for all approaches.

\begin{table*}[t]
\centering
\caption{Detailed benchmark results across six stress-test environments. For SRCC, higher is better. For discovery latency, lower is better. Values are means over 20 independent trials.}
\label{tab:appendix_six_env_results}
\small
\resizebox{\textwidth}{!}{%
\begin{tabular}{llcccccc}
\toprule
Environment & Method & SRCC@500 $\uparrow$ & SRCC@1500 $\uparrow$ & SRCC@3000 $\uparrow$ & Discovery $\downarrow$ & Discovery Success $\uparrow$ & Top-1 After Injection $\uparrow$ \\
\midrule
\multirow{9}{*}{\texttt{reference}}
& Elo                  & 0.5440 & 0.5541 & 0.5339 & 41.00 & 0.40 & 0.15 \\
& TrueSkill            & 0.4079 & 0.5005 & 0.5610 & 6.20  & 1.00 & 0.50 \\
& TrueSkill2           & 0.4523 & 0.5968 & 0.6521 & 6.65  & 1.00 & 0.50 \\
& Glicko-2             & 0.4074 & 0.7078 & 0.8097 & 20.60 & 0.90 & 0.25 \\
& Rank Centrality      & 0.4276 & 0.5660 & 0.6471 & 40.40 & 0.25 & 0.25 \\
& K-Sort Arena         & 0.7474 & 0.8512 & 0.9283 & 27.20 & 1.00 & 0.35 \\
& Active Ranking (AR)  & 0.3161 & 0.4633 & 0.5608 & 2.75  & 1.00 & 0.90 \\
& RUCB                 & 0.6606 & 0.7602 & 0.7779 & 6.45  & 1.00 & 0.75 \\
& Ours                 & \textbf{0.8293} & \textbf{0.8845} & \textbf{0.9546} & 5.75 & \textbf{1.00} & 0.55 \\
\midrule
\multirow{9}{*}{\texttt{noisier\_judge}}
& Elo                  & 0.4645 & 0.6136 & 0.4620 & 41.10 & 0.50 & 0.15 \\
& TrueSkill            & 0.4241 & 0.5202 & 0.6035 & 6.05  & 1.00 & 0.50 \\
& TrueSkill2           & 0.4153 & 0.5766 & 0.6411 & 6.15  & 1.00 & 0.50 \\
& Glicko-2             & 0.3738 & 0.7279 & 0.7977 & 20.45 & 0.90 & 0.25 \\
& Rank Centrality      & 0.4307 & 0.5275 & 0.6697 & 40.55 & 0.25 & 0.25 \\
& K-Sort Arena         & 0.6642 & 0.8358 & \textbf{0.9020} & 26.75 & \textbf{1.00} & 0.40 \\
& Active Ranking (AR)  & 0.3123 & 0.4241 & 0.5648 & \textbf{2.75}  & \textbf{1.00} & \textbf{0.90} \\
& RUCB                 & 0.6883 & 0.7446 & 0.8042 & 10.95 & 0.95 & 0.65 \\
& Ours                 & \textbf{0.8236} & \textbf{0.9007} & 0.8912 & 5.75 & \textbf{1.00} & 0.55 \\
\midrule
\multirow{9}{*}{\texttt{hard\_extreme}}
& Elo                  & 0.4119 & 0.6193 & 0.3964 & 34.85 & 0.80 & 0.15 \\
& TrueSkill            & 0.4141 & 0.5259 & 0.5455 & 9.20  & 0.95 & 0.40 \\
& TrueSkill2           & 0.3922 & 0.5417 & 0.5629 & 6.85  & 1.00 & 0.45 \\
& Glicko-2             & 0.3895 & 0.6904 & 0.8076 & 16.50 & 0.95 & 0.20 \\
& Rank Centrality      & 0.3946 & 0.6095 & 0.6808 & 44.25 & 0.25 & 0.30 \\
& K-Sort Arena         & 0.6263 & 0.8064 & \textbf{0.9366} & 26.15 & 1.00 & 0.35 \\
& Active Ranking (AR)  & 0.3195 & 0.4521 & 0.5410 & \textbf{1.00}  & 1.00 & \textbf{0.95} \\
& RUCB                 & 0.6974 & 0.8251 & 0.8563 & 8.30  & 0.90 & 0.65 \\
& Ours                 & \textbf{0.8082} & \textbf{0.8809} & 0.8937 & 5.70 & \textbf{1.00} & 0.50 \\
\midrule
\multirow{9}{*}{\texttt{less\_extreme}}
& Elo                  & 0.4260 & 0.6469 & 0.5422 & 36.90 & 0.60 & 0.15 \\
& TrueSkill            & 0.4533 & 0.5580 & 0.6212 & 6.30  & 1.00 & 0.55 \\
& TrueSkill2           & 0.4924 & 0.6390 & 0.6964 & 6.35  & 1.00 & 0.55 \\
& Glicko-2             & 0.2555 & 0.6414 & 0.7831 & 18.05 & 0.90 & 0.15 \\
& Rank Centrality      & 0.3805 & 0.5630 & 0.6397 & 44.75 & 0.25 & 0.25 \\
& K-Sort Arena         & 0.6798 & 0.9251 & \textbf{0.9240} & 26.15 & 1.00 & 0.35 \\
& Active Ranking (AR)  & 0.3262 & 0.4208 & 0.5108 & \textbf{5.35}  & 0.95 & \textbf{0.90} \\
& RUCB                 & 0.6448 & 0.7527 & 0.7673 & 11.90 & 0.95 & 0.65 \\
& Ours                 & \textbf{0.8558} & \textbf{0.9376} & 0.8903 & 5.90 & \textbf{1.00} & 0.50 \\
\midrule
\multirow{9}{*}{\texttt{prompt\_calibrated}}
& Elo                  & 0.4541 & 0.5740 & 0.5464 & 39.25 & 0.55 & 0.15 \\
& TrueSkill            & 0.3917 & 0.5554 & 0.4984 & 6.10  & 1.00 & 0.50 \\
& TrueSkill2           & 0.4412 & 0.5991 & 0.6691 & 6.30  & 1.00 & 0.50 \\
& Glicko-2             & 0.4266 & 0.6809 & 0.8231 & 20.15 & 0.90 & 0.25 \\
& Rank Centrality      & 0.4336 & 0.5898 & 0.6819 & 38.50 & 0.30 & 0.25 \\
& K-Sort Arena         & 0.6558 & 0.8477 & 0.8877 & 26.30 & 1.00 & 0.35 \\
& Active Ranking (AR)  & 0.3356 & 0.4548 & 0.5284 & \textbf{2.90}  & 1.00 & \textbf{0.90} \\
& RUCB                 & 0.7335 & 0.7823 & 0.8327 & 9.60  & 0.90 & 0.65 \\
& Ours                 & \textbf{0.7970} & \textbf{0.9054} & \textbf{0.9104} & 5.80 & \textbf{1.00} & 0.55 \\
\midrule
\multirow{9}{*}{\texttt{prompt\_clustered\_stress}}
& Elo                  & 0.3715 & 0.4314 & 0.5980 & 36.05 & 0.65 & 0.20 \\
& TrueSkill            & 0.3732 & 0.4476 & 0.4936 & 7.15  & 1.00 & 0.40 \\
& TrueSkill2           & 0.3688 & 0.4789 & 0.5079 & 6.50  & 1.00 & 0.40 \\
& Glicko-2             & 0.3819 & 0.5457 & 0.5872 & 21.20 & 0.95 & 0.35 \\
& Rank Centrality      & 0.3095 & 0.4321 & 0.4895 & 44.05 & 0.25 & 0.25 \\
& K-Sort Arena         & 0.5746 & 0.5944 & \textbf{0.6300} & 26.45 & 1.00 & 0.40 \\
& Active Ranking (AR)  & 0.3665 & 0.4602 & 0.5338 & \textbf{1.05}  & 1.00 & \textbf{1.00} \\
& RUCB                 & 0.4875 & 0.5970 & 0.6212 & 13.90 & 0.90 & 0.60 \\
& Ours                 & \textbf{0.6029} & \textbf{0.6141} & 0.5969 & 6.05 & \textbf{1.00} & 0.55 \\
\bottomrule
\end{tabular}%
}
\end{table*}

\paragraph{A closer reading of the trade-offs.}
The results clarify that different methods optimize different objectives.

\textsc{Active Ranking (AR)} is the strongest method if the sole goal is to identify the newly injected best model as quickly as possible. Its discovery latency is consistently the smallest, and its top-1-after-injection rate is often the highest. However, its global rank correlation is poor in every environment, which makes it unsuitable when the goal is to maintain a reliable leaderboard rather than only detect the best arm.

\textsc{K-Sort Arena} is the strongest baseline in terms of eventual ranking precision. In several environments, it achieves the highest SRCC@3000 and remains a highly competitive reference for methods optimized toward asymptotic rank recovery. Its weakness is responsiveness: it is consistently much slower at detecting newly injected strong models, which is a serious limitation for dynamic evaluation settings where leaderboards must react quickly to new entrants.

\textsc{RUCB} offers a more balanced baseline and remains reasonably competitive under several forms of stochastic stress. Nevertheless, its performance is less stable than \textsc{Ours}, especially under prompt-level heterogeneity and clustered prompt bias, where both its discovery latency and rank correlation deteriorate.

Traditional rating systems such as \textsc{TrueSkill}, \textsc{TrueSkill2}, and \textsc{Glicko-2} can still produce acceptable final rankings in some milder settings, but they are generally less robust to the full spectrum of dynamic perturbations considered here. \textsc{Elo} and \textsc{Rank Centrality} are consistently dominated and are poorly suited to dynamic model injection.

\paragraph{Takeaway.}
Overall, this extended analysis supports the same conclusion as the main paper, but in a more fine-grained way. Our method is not always the absolute best on the most asymptotic metric, but it is the most consistent method in the regime that matters for dynamic leaderboards: it produces the strongest early-stage rankings, maintains near-constant and low discovery latency, and preserves perfect discovery success across all six stress-test environments. This combination makes it a particularly strong choice when both ranking quality and responsiveness must be optimized under limited comparison budgets.

\begin{figure*}[t]
    \centering
    \includegraphics[width=0.9\textwidth]{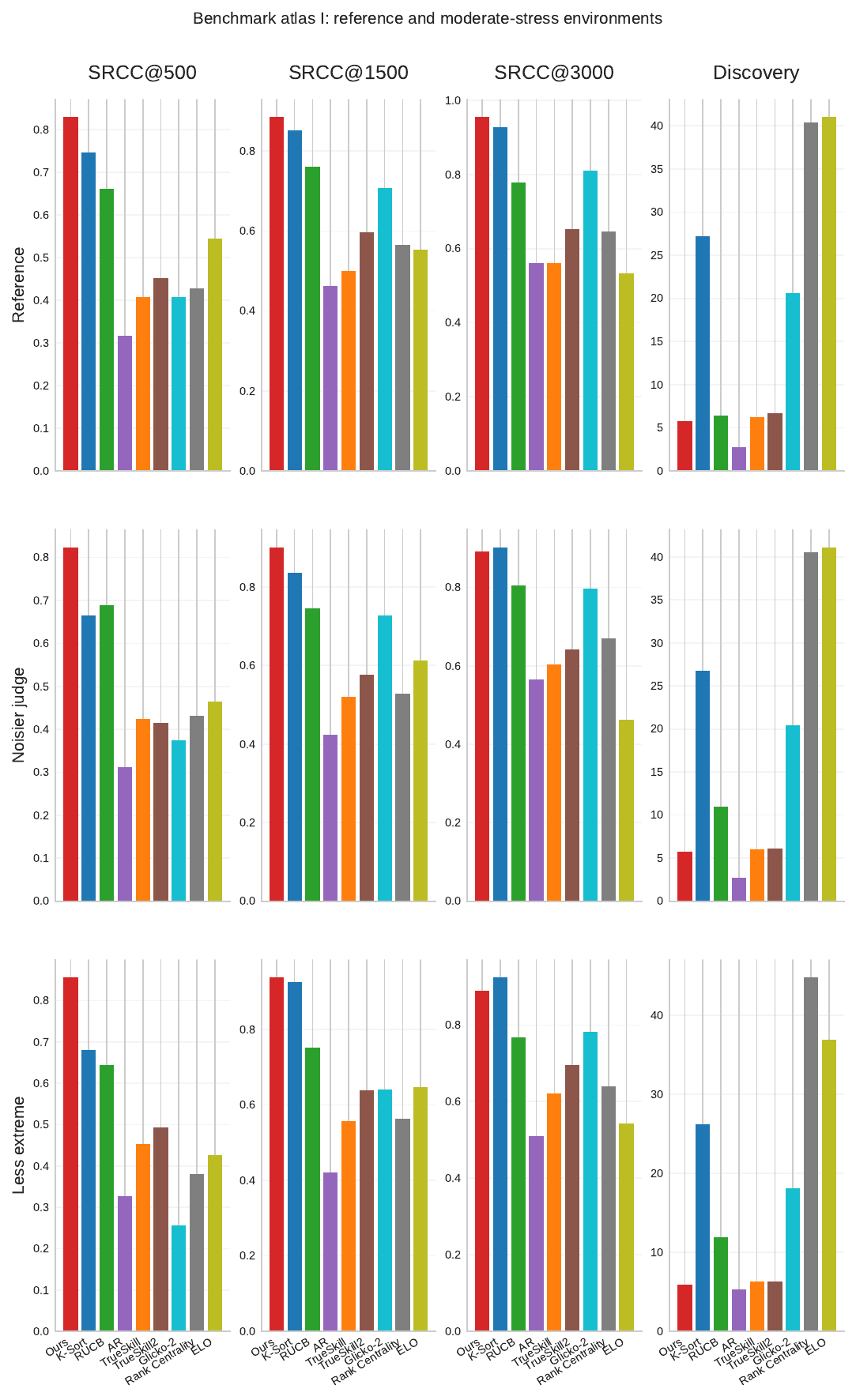}
    \caption{Benchmark atlas I covering the reference and moderate-stress settings: \texttt{reference}, \texttt{noisier\_judge}, and \texttt{less\_extreme}. Each row corresponds to one environment and each column corresponds to one key metric: SRCC@500, SRCC@1500, SRCC@3000, and late-entry discovery. This split layout avoids the over-compression of a single all-in-one page while preserving direct all-baseline comparison.}
    \label{fig:appendix_six_env_method_bars_part1}
\end{figure*}

\begin{figure*}[t]
    \centering
    \includegraphics[width=0.9\textwidth]{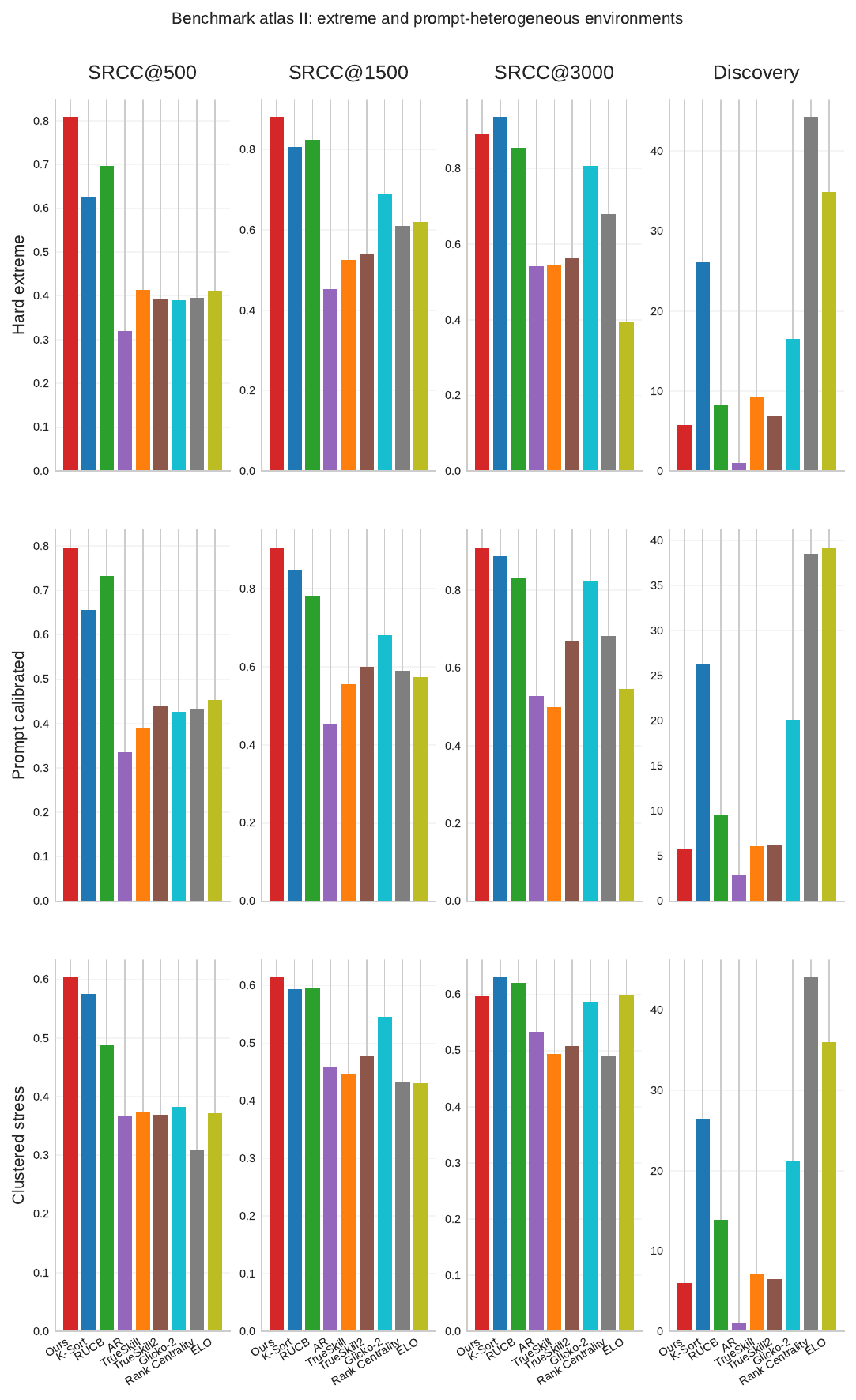}
    \caption{Benchmark atlas II covering the more difficult settings: \texttt{hard\_extreme}, \texttt{prompt\_calibrated}, and \texttt{prompt\_clustered\_stress}. These environments stress extreme-event behavior and prompt-level heterogeneity more strongly. The split view makes the comparative pattern easier to read and highlights that our method remains especially strong in early-stage ranking quality under the hardest conditions.}
    \label{fig:appendix_six_env_method_bars_part2}
\end{figure*}

\begin{figure*}[t]
    \centering
    \includegraphics[width=0.92\textwidth]{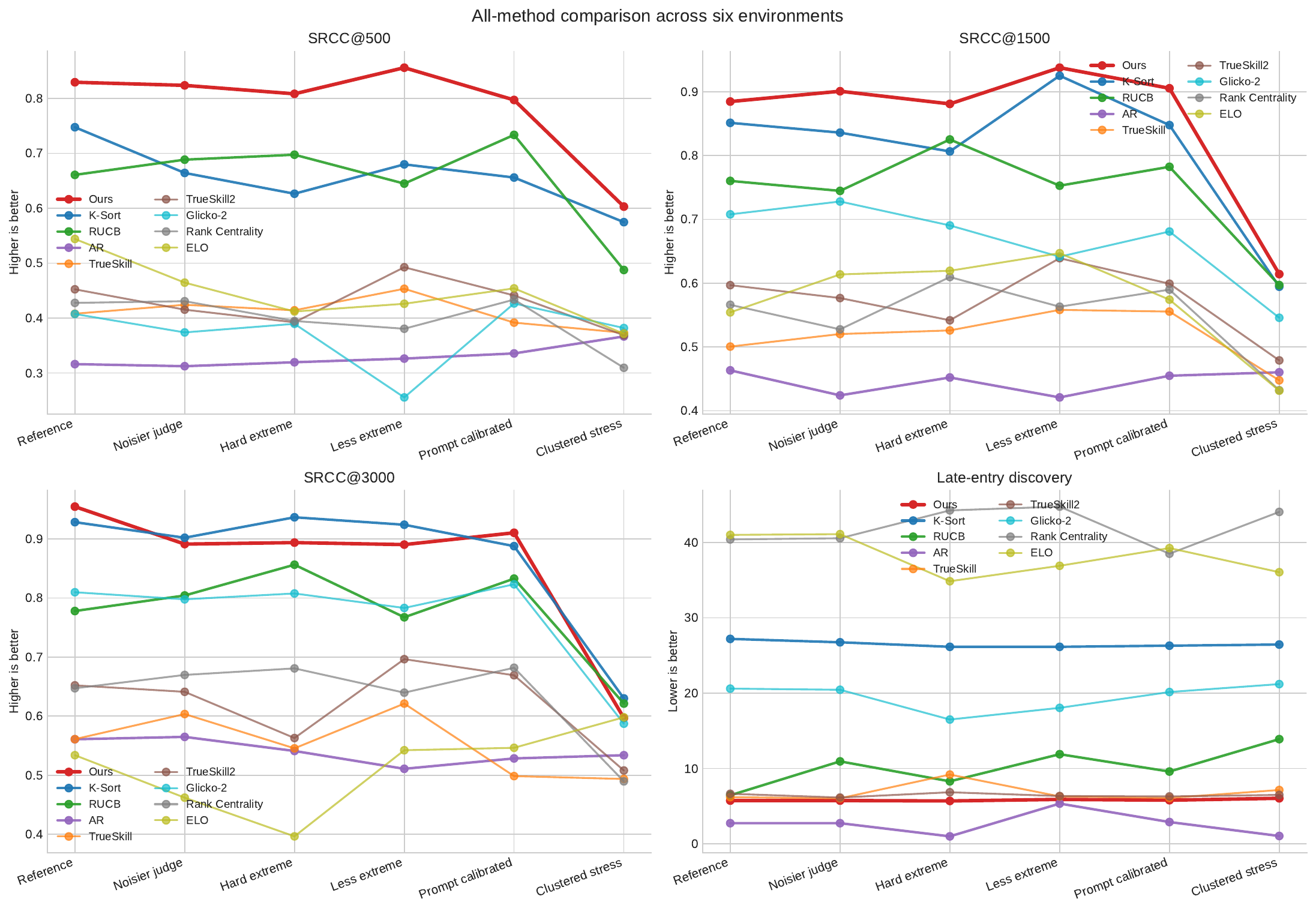}
    \caption{All-method comparison across six environments shown as line plots. Compared with the tables, the line view highlights cross-environment stability more directly. Our method remains uniformly strong on SRCC@500 and SRCC@1500, while K-Sort Arena and AR emphasize different ends of the long-run-fidelity versus rapid-discovery trade-off.}
    \label{fig:appendix_all_methods_line_grid}
\end{figure*}

\begin{figure*}[t]
    \centering
    \includegraphics[width=0.92\textwidth]{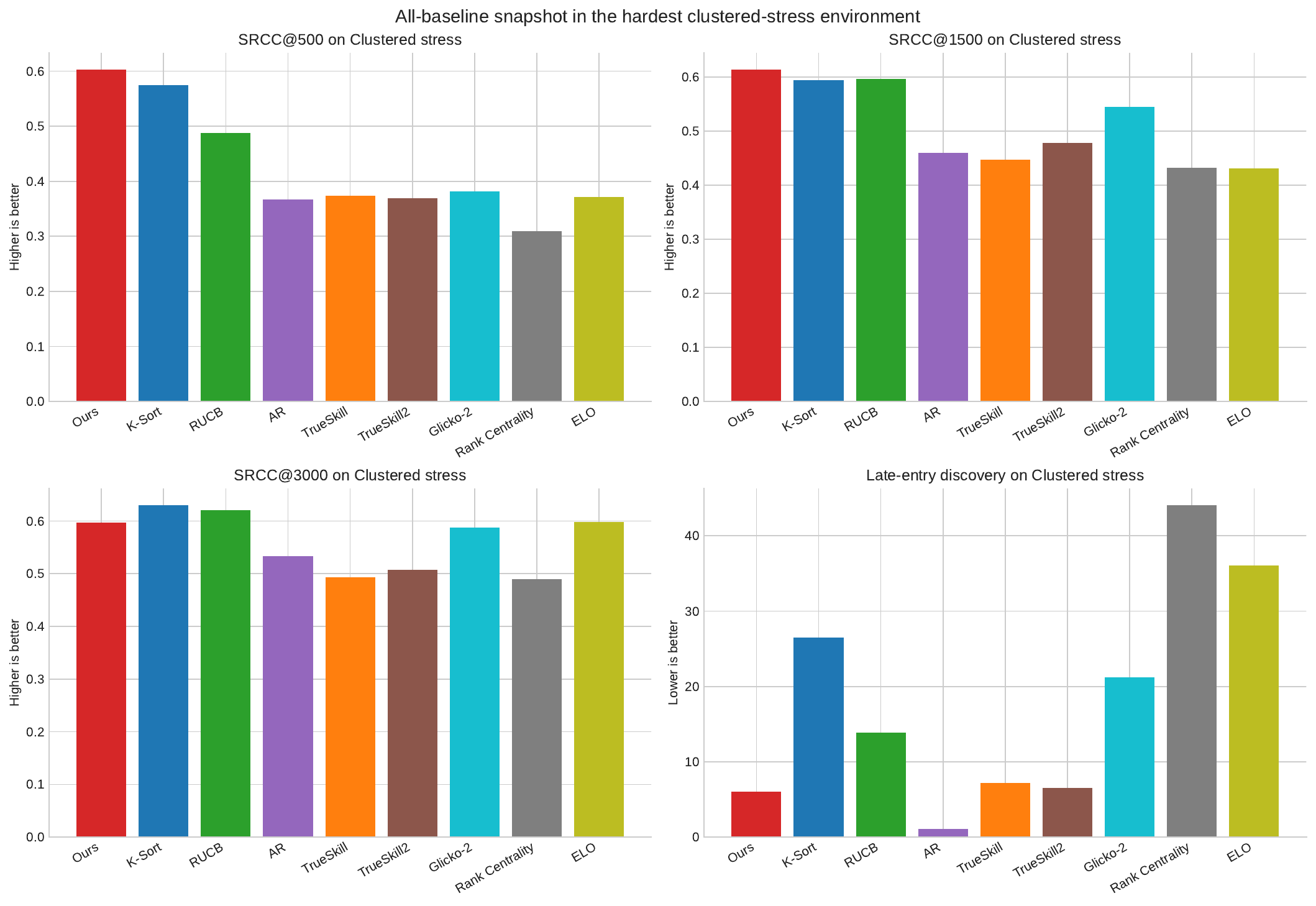}
    \caption{Expanded comparison in the hardest environment, \texttt{prompt\_clustered\_stress}. This setting is particularly difficult because prompt clusters induce structured model-dependent bias. Even in this case, our method still gives the best early-stage SRCC, while K-Sort Arena retains the strongest long-run SRCC and Active Ranking remains the fastest discovery specialist.}
    \label{fig:appendix_clustered_stress_bars}
\end{figure*}

\begin{figure*}[t]
    \centering
    \includegraphics[width=\textwidth]{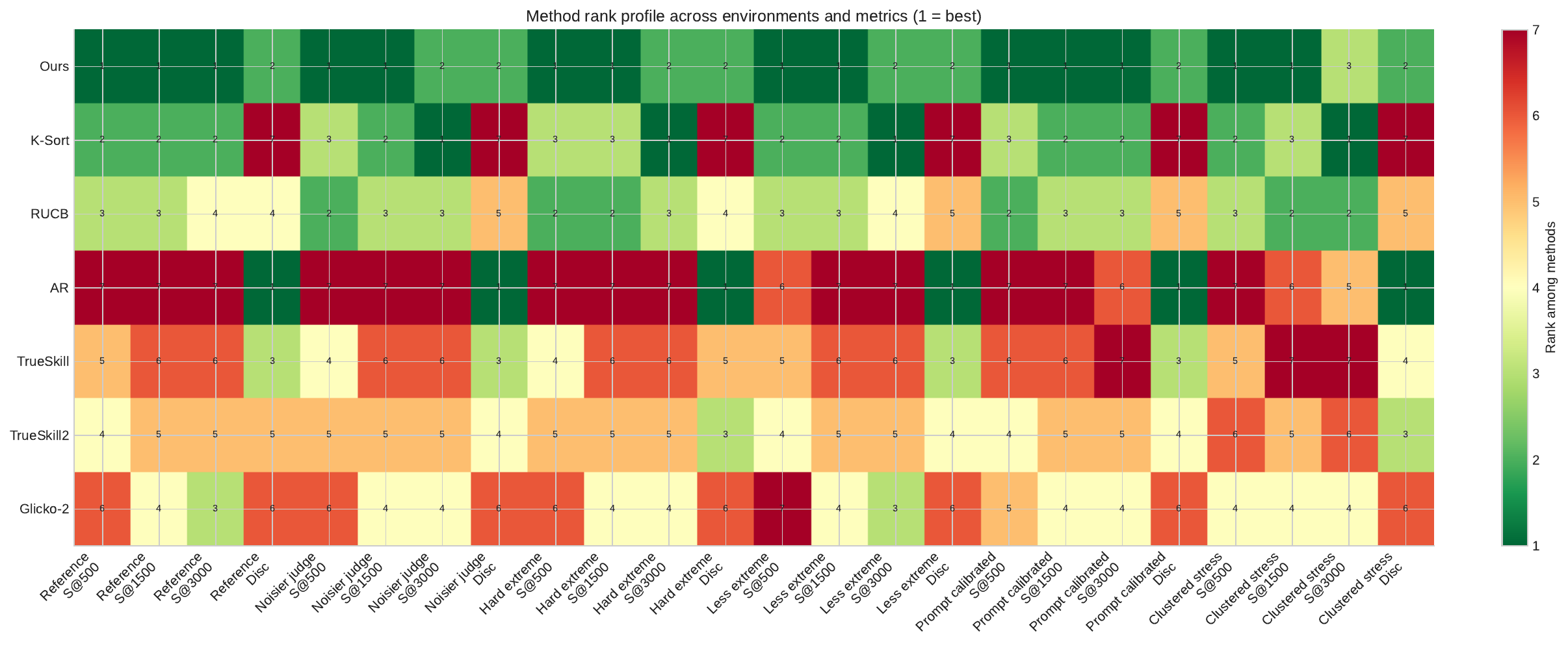}
    \caption{Rank profile heatmap across environments and metrics. Each cell shows the rank of a method under one environment--metric pair, with lower ranks being better. The heatmap provides a compact summary of consistency across the full robustness suite and visually confirms that our method dominates the early-stage ranking regime more systematically than the baselines.}
    \label{fig:appendix_method_rank_heatmap}
\end{figure*}

\begin{figure*}[t]
    \centering
    \includegraphics[width=\textwidth]{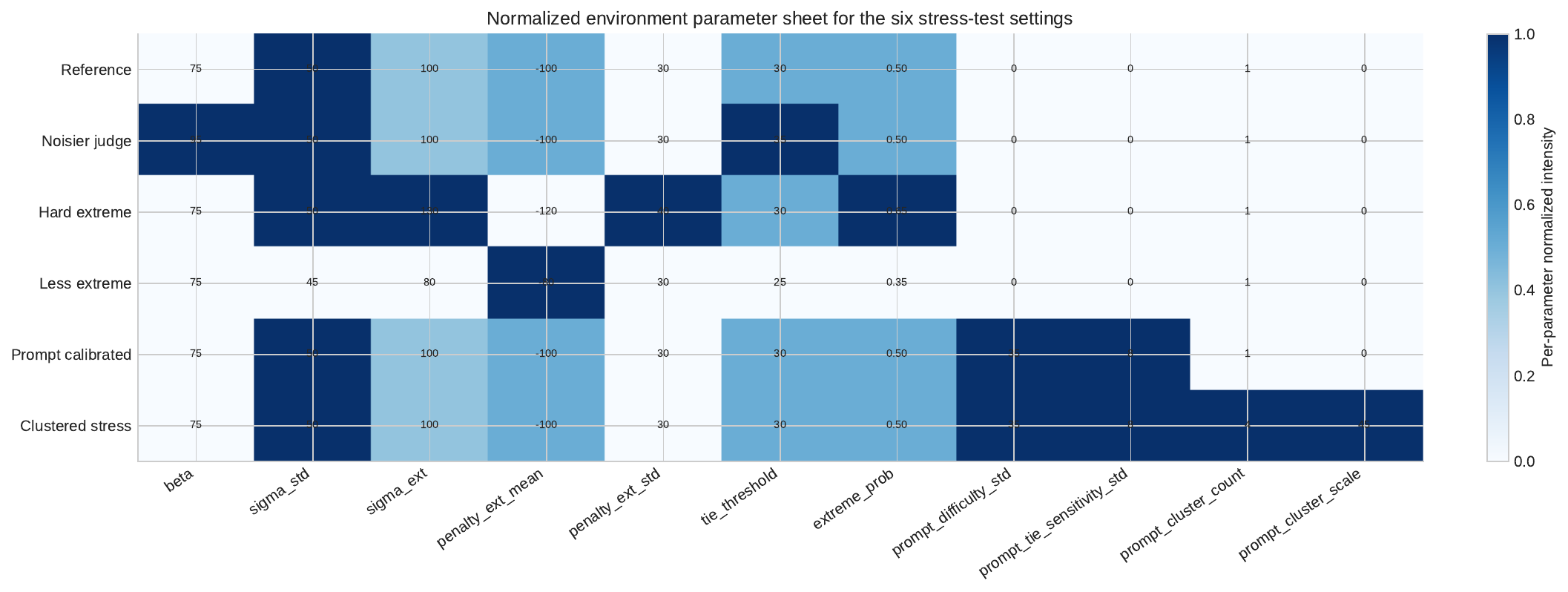}
    \caption{Parameter sheet for the six simulated environments. The figure visualizes how the stress-test suite varies evaluator noise, extreme-event severity, tie tendency, and prompt-level heterogeneity. This complements Table~\ref{tab:appendix_env_definitions} by giving a compact visual overview of the full robustness design space.}
    \label{fig:appendix_env_parameter_sheet}
\end{figure*}

\subsection{Figure-based Interpretation of the Six-Environment Results}

To complement the tables above, we further visualize the robustness study from several angles. These figures are not intended to introduce new claims beyond Table~\ref{tab:appendix_six_env_results}; rather, they help make the underlying trade-offs easier to read at a glance and clarify why the proposed method is favorable for dynamic leaderboards.

Figures~\ref{fig:appendix_six_env_method_bars_part1} and~\ref{fig:appendix_six_env_method_bars_part2} provide the broadest view. By placing all methods, all six environments, and four key metrics into a single atlas, it makes two patterns immediately visible. First, our method is the most consistently strong method in the \emph{early-stage} ranking regime: across environments, its bars for SRCC@500 and SRCC@1500 remain near the top with much smaller variation than the baselines. Second, different baselines are revealed as specialists rather than uniformly strong competitors. \textsc{K-Sort Arena} is strong on SRCC@3000 but weak on discovery latency, while \textsc{Active Ranking} is excellent on late-entry discovery but much weaker on global ranking fidelity. This is exactly the trade-off we emphasize in the main paper: dynamic evaluation should reward methods that balance responsiveness and ranking quality, rather than optimize one objective in isolation.

Figure~\ref{fig:appendix_all_methods_line_grid} makes the same point from a cross-environment perspective. Unlike the bar atlas, the line view emphasizes \emph{stability} as the environment changes. Our method traces a relatively flat and high trajectory on SRCC@500 and SRCC@1500, indicating that the gain is not tied to one particular environment design. By contrast, several baselines fluctuate more substantially as evaluator noise, extreme events, or prompt-level heterogeneity become more severe. This matters for real-world dynamic benchmarking, where the evaluation process is rarely stationary: the most useful scheduler is not merely the one that wins in a clean reference environment, but the one that degrades gracefully when the comparison process becomes noisier or more heterogeneous.

Figure~\ref{fig:appendix_clustered_stress_bars} zooms in on the hardest setting, \texttt{prompt\_clustered\_stress}. We highlight this environment separately because it is the most adversarial one in the suite: prompts are not only heterogeneous, but clustered in a way that can systematically favor or penalize subsets of models. In this regime, all methods degrade relative to the easier environments, which is itself an informative result. Nevertheless, our method still attains the strongest SRCC@500 and SRCC@1500. This point is important for our narrative: even when final asymptotic recovery becomes difficult for everyone, our scheduler still produces the most useful \emph{early} leaderboard, which is often the practically relevant regime in a continually expanding benchmark.

Figure~\ref{fig:appendix_method_rank_heatmap} offers a compact summary of consistency. Instead of focusing on raw metric magnitudes, it converts each environment--metric pair into a within-group rank and visualizes the result as a heatmap. This removes some of the distraction caused by absolute scale differences across metrics and highlights a more structural fact: our method occupies top ranks much more often in the early-stage cells than the baselines. In other words, the advantage is not confined to a small number of cherry-picked conditions. The heatmap also makes clear that the leading baselines occupy different corners of the design space: \textsc{K-Sort Arena} wins some late-stage ranking cells, \textsc{Active Ranking} wins discovery-specialist cells, and \textsc{RUCB} is often competitive but less stable under stress. This visual decomposition is favorable to our core claim because it shows that our method is the strongest \emph{generalist} in the dynamic regime.

Finally, Figure~\ref{fig:appendix_env_parameter_sheet} documents the environment suite itself. Its role is methodological rather than competitive: it shows that the six environments vary along multiple axes, including evaluator noise, extreme-event severity, tie behavior, and prompt-level perturbation. This strengthens the credibility of the robustness study because the improvements of our method are not being demonstrated on six near-duplicates of the same setting. Instead, the suite spans a reasonably diverse set of stressors, and the figures above show that the proposed method remains favorable across that broader design space.

Taken together, these visualizations reinforce the main interpretation of the appendix tables. The proposed method is not defined by being the single best point on one asymptotic metric. Its strength is that it repeatedly delivers the best or near-best \emph{early} global ordering, while keeping late-entry discovery fast and highly stable. For dynamic leaderboards, this is the most practically relevant operating regime, because rankings are consumed while evaluation is still ongoing and new models may appear after the leaderboard has already begun to stabilize.

%\clearpage
%\input{checklist.tex}
\end{document}